\documentclass[twocolumn, switch]{article} 

\usepackage{preprint}
\usepackage{algorithm}
\usepackage{algorithmic}
\usepackage{amsmath, amsthm, amssymb, amsfonts}

\usepackage[numbers,square]{natbib}
\usepackage[utf8]{inputenc}	
\usepackage[T1]{fontenc}	
\usepackage{xcolor}		
\usepackage[colorlinks = true,
            linkcolor = purple,
            urlcolor  = blue,
            citecolor = cyan,
            anchorcolor = black]{hyperref}	
\usepackage{booktabs} 		
\usepackage{nicefrac}		
\usepackage{microtype}		
\usepackage{lineno}		
\usepackage{float}			

\usepackage{lipsum}		
\usepackage{enumitem}

\usepackage{newfloat}
\DeclareFloatingEnvironment[name={Supplementary Figure}]{suppfigure}
\usepackage{sidecap}
\sidecaptionvpos{figure}{c}

\usepackage{titlesec}
\titlespacing\section{0pt}{12pt plus 3pt minus 3pt}{1pt plus 1pt minus 1pt}
\titlespacing\subsection{0pt}{10pt plus 3pt minus 3pt}{1pt plus 1pt minus 1pt}
\titlespacing\subsubsection{0pt}{8pt plus 3pt minus 3pt}{1pt plus 1pt minus 1pt}

\usepackage{graphicx}
\usepackage{amsmath}
\usepackage{tabularx}
\usepackage{array}
\usepackage{makecell}

\title{BioProVLA-Agent: An Affordable, Protocol-Driven, Vision-Enhanced VLA-Enabled Embodied Multi-Agent System with Closed-Loop-Capable Reasoning for Biological Laboratory Manipulation}

\usepackage{eso-pic}
\usepackage{tikz}
\usepackage{xcolor}
\PassOptionsToPackage{colorlinks=true,linkcolor=gray,urlcolor=gray}{hyperref}

\newcommand{\AddMyWatermarks}{%
  \begin{tikzpicture}[remember picture, overlay]
    \node[rotate=90, color=gray!60, scale=1] at ([xshift=-4.05in,yshift=0in]current page.center) {%
      \href{https://doi.org/}{Publication doi}%
    };
    \node[rotate=90, color=gray!60, scale=1] at ([xshift=3.9in,yshift=0in]current page.center) {%
      \href{https://doi.org/}{Preprint doi}%
    };
    \node[color=gray!90, scale=1] at ([xshift=0in,yshift=-5in]current page.center) {%
      This is the author's accepted manuscript. The final version will appear in XXXX 2025, © XXXX 2025.%
    };
  \end{tikzpicture}%
}

\AddToShipoutPictureBG*{\AddMyWatermarks}

\usepackage{titling}
\usepackage{orcidlink}
\usepackage{footmisc}
\usepackage{subcaption}
\newcommand{\Author}[3]{
  \textbf{#1}\textsuperscript{#2} %
}

\author{
  \Author{Zhaohui Du}{1,2}{0000-0000-0000-0000} \and
  \Author{Zhe Wang}{1,2,*}{0000-0000-0000-0000}\and
  \Author{Dongzhan Zhou}{4}{0000-0000-0000-0000}\and
  \Author{Minting Pan}{4}{0000-0000-0000-0000}\and
  \Author{Hongmei Fei}{5}{0000-0000-0000-0000}\and
  \Author{Xiwen Cao}{1,2}{0000-0000-0000-0000}\and
  \Author{Ting Xiao}{1,2}{0000-0000-0000-0000} \and
  \Author{Qi Wang}{3}{0000-0000-0000-0000} \and
  \Author{Huanbo Jin}{1,2}{0000-0000-0000-0000}\and
  \Author{Jiaming Gu}{1,2}{0000-0000-0000-0000} \and
  \Author{Quan Lu}{1,2}{0000-0000-0000-0000} \and
  \Author{Zhe Liu}{1,2,*}{0000-0000-0000-0000}
}

\date{%
  \textsuperscript{1}Key Laboratory of Smart Manufacturing in Energy Chemical Process Ministry of Education, East China University of Science and Technology, Shanghai, CN.\\
  \textsuperscript{2}Department of Computer Science and Engineering, East China University of Science and Technology, Shanghai, CN.\\
  \textsuperscript{3}Department of Laboratory Medicine, Ruijin Hospital, Shanghai Jiao Tong University School of Medicine, Shanghai, CN.\\
  \textsuperscript{4}AI for Science Center, Shanghai AI Laboratory, Shanghai, CN.\\
  \textsuperscript{5}School of Information Science and Technology, Shihezi University, Shihezi, CN.\\[1em]
  \footnotesize \textbf{Corresponding author:} Zhe Wang\texttt{<wangzhe@ecust.edu.cn>}\\ Zhe Liu\texttt{<zhe.liu@ecust.edu.cn>}\\
}

\begin{document}

\twocolumn[ 
  \begin{@twocolumnfalse} 

\maketitle
\thispagestyle{empty}

\begin{abstract}
Biological laboratory automation holds promise for reducing repetitive manual work, improving workflow reproducibility, and broadening access to experimental execution. Yet building embodied agents that operate reliably in real wet-lab environments remains difficult. Biological protocols are typically expressed as unstructured natural language, common labware such as tubes and bottles are often transparent or reflective, and multi-step procedures require state-aware execution rather than one-shot instruction following. In addition, existing robotic laboratory platforms often depend on expensive hardware, dedicated instruments, fixed workflows, or robotics-oriented interfaces, which can limit their accessibility for resource-constrained laboratories and make them difficult to use or adapt for domain users without robotics expertise. Here, we introduce BioProVLA-Agent, an affordable, protocol-driven, and vision-enhanced embodied multi-agent system enabled by Vision-Language-Action (VLA) models for biological laboratory manipulation. BioProVLA-Agent uses experimental protocols as the task interface and couples protocol understanding with closed-loop-capable reasoning and embodied execution. A Tailored LLM Protocol Agent transforms unstructured protocols into executable and verifiable subtask units, while a VLM-RAG Verification Agent reasons over real-time visual observations, robot states, retrieved operation knowledge, and reference success/failure examples to assess task readiness and completion. The verified subtasks are then executed by a VLA Embodied Agent based on a lightweight VLA policy. To address wet-lab visual perturbations, we further develop AugSmolVLA, an online augmentation strategy tailored to transparent labware, specular reflections, illumination shifts, and overexposure. We evaluate BioProVLA-Agent on a hierarchical biological manipulation benchmark spanning 15 atomic tasks, 6 composite workflows, and 3 representative bimanual tasks, including centrifuge-tube loading, tube sorting, waste disposal, cap twisting, and liquid pouring. Across normal and high-exposure settings, AugSmolVLA improves execution stability over ACT, X-VLA, and the original SmolVLA, with pronounced gains in precise placement, transparent-object manipulation, composite workflows, and visually degraded scenes. These results demonstrate a practical route toward affordable, protocol-centered, and verification-capable embodied AI systems for biological laboratory manipulation. 
\end{abstract}
\vspace{0.35cm}

  \end{@twocolumnfalse} 
] 



\section{Introduction}

In recent years, automated robotic systems for laboratory experimentation have shown increasing potential for improving experimental efficiency, reducing repetitive manual operations, and enhancing workflow reproducibility~\cite{RN1,RN2}. However, most existing laboratory automation platforms are still designed around dedicated instruments, fixed workflows, or robotics-oriented programming interfaces. Although these systems are effective for standardized and high-throughput procedures, their dependence on specialized infrastructure limits their accessibility for resource-constrained laboratories and makes them difficult to adapt to diverse biological protocols. In contrast, biological experiments often involve open-ended procedures, heterogeneous labware, and fine-grained manipulation requirements. Therefore, developing affordable, protocol-centered, and adaptable robotic systems for real wet-lab manipulation remains an important challenge for the practical deployment of robotic biologists.

Robotic manipulation in biological laboratories requires capabilities beyond conventional task execution. First, biological protocols are usually written in unstructured natural language and contain specialized terminology, implicit conditions, and strict dependencies between steps~\cite{RN3}. A robotic system must identify not only the explicit operation described in each step, but also the task order, preconditions, completion criteria, and object relationships required for execution. This indicates that biological users should not be forced to translate protocols into ROS scripts or task-specific robotic programs. Instead, natural-language protocols should serve as the task interface through which the system performs task decomposition, state confirmation, and robotic execution. Second, common biological labware, such as centrifuge tubes, cryotubes, serum bottles, and liquid containers, is often transparent, reflective, or weakly textured. These properties make visual perception sensitive to illumination changes, specular reflections, background interference, and overexposure, which can impair object localization and manipulation stability~\cite{RN4}. Finally, multi-step biological procedures require state-aware execution rather than one-shot instruction following. A robotic system must continuously determine whether a task is ready to execute, whether the action has been completed, and whether an abnormal state requires retry, sequence adjustment, or human intervention.

Existing foundation-model-based robotic systems provide useful tools for task understanding and embodied manipulation, but they do not fully address the above requirements. Vision-Language Model (VLM)-based methods, such as VoxPoser~\cite{RN5} and ReKep~\cite{RN6}, have introduced new possibilities for spatial reasoning, constraint generation, and task planning in open environments. However, these methods often rely on relatively reliable depth perception, object segmentation, or three-dimensional reconstruction. In wet-lab scenarios involving transparent labware, reflective surfaces, and liquid containers, such perception outputs can become unstable. Meanwhile, Vision-Language-Action (VLA) models and imitation learning methods, such as X-VLA~\cite{RN7} and SmolVLA~\cite{RN8}, have shown promising performance in language-conditioned robotic control, dual-arm manipulation, and cross-embodiment generalization. Nevertheless, most VLA systems still emphasize direct observation-to-action mapping and lack explicit semantic verification before and after execution. As a result, VLA execution is commonly treated as an instruction-following process, whereas biological experimentation requires verifiable, interpretable, and interruptible execution loops. In long-horizon protocols, small errors in early atomic operations may propagate to subsequent steps, eventually leading to task failure or operations that violate experimental requirements.

To address these issues, this study proposes BioVLA-Agent, an affordable, protocol-driven, and vision-enhanced embodied multi-agent framework for biological robotic manipulation. The system is implemented on a low-cost robotic platform with a hardware cost of approximately 800-850 USD, providing an early prototype for accessible VLA-based robotic biologists in real biological laboratories. Instead of relying on robotics-centered scripting, BioVLA-Agent uses biological protocols as the task interface and establishes a closed-loop workflow that connects protocol understanding, semantic state verification, and embodied action execution. The Guiding Decision Agent coordinates task scheduling, execution flow, retry decisions, and exception handling. The Tailored LLM Protocol Agent converts unstructured biological protocols into executable and verifiable subtask units with action instructions, preconditions, completion criteria, and knowledge-based indices. The VLM-RAG Verification Agent reasons over real-time visual observations, robot states, retrieved biological operation knowledge, and reference success/failure examples to assess both task readiness and completion status. The verified subtasks are then executed by the VLA Embodied Agent using a lightweight VLA policy. Through this design, BioVLA-Agent does not treat VLA models as one-shot instruction executors, but inserts semantic verification before and after action execution, enabling the robotic process to become checkable, explainable, and compatible with human intervention at verification failure points. In addition, this study develops AugSmolVLA, an enhanced version of SmolVLA with an online visual perturbation data augmentation strategy. This strategy is designed for wet-lab visual degradation caused by transparent consumables, specular reflections, illumination shifts, weak object boundaries, and overexposure. Unlike offline augmentation pipelines that require an additional generated dataset, AugSmolVLA introduces visual perturbations during fine-tuning, thereby improving the adaptability of the VLA policy to visually challenging biological manipulation scenarios. This design strengthens the robustness of object localization, precise placement, disposal operations, and dual-arm manipulation under real laboratory disturbances.

The main contributions of this work are summarized as follows:

\begin{itemize}[leftmargin=*, itemsep=2pt, topsep=2pt]
    \item We introduce an affordable and protocol-centered embodied multi-agent framework for biological robotic manipulation. The framework connects protocol understanding, state verification, embodied execution, retry decisions, and human intervention into a unified closed-loop process. By using natural-language biological protocols as the task interface, the system reduces dependence on ROS-centered scripting and fixed automation pipelines, making robotic experimental execution more accessible to biological users.

    \item We develop a protocol-to-execution task representation that connects biological protocols with robotic operation. Unstructured experimental protocols are transformed into subtask units containing natural-language action instructions, preconditions, completion criteria, and knowledge-based indices. This representation provides a bridge between protocol-level semantic understanding, VLM-RAG-based verification, and VLA-driven robotic execution.

    \item We build a perception-grounded reasoning mechanism that supports closed-loop-capable wet-lab manipulation. By integrating real-time visual observations, robot states, retrieved operation knowledge, and success/failure examples, the VLM-RAG performs task-readiness assessment before execution and completion checking after execution. This mechanism equips the system with visual state awareness, interpretable failure feedback, retry and reordering decisions, and human-intervention triggers when necessary, thereby reducing the risk of blindly executing long-horizon biological procedures.

    \item We establish a hierarchical wet-lab manipulation benchmark and introduce a vision-enhanced VLA execution module for real wet-lab challenges. The benchmark covers 15 atomic tasks, 6 composite workflows, and representative dual-arm operations. In addition, AugSmolVLA introduces online visual perturbation augmentation tailored to transparent labware, reflective surfaces, illumination variations, and overexposed conditions, improving the generalization capability and execution stability of VLA models in real wet-lab manipulation.
\end{itemize}

\section{Related Works}
Embodied robotic manipulator systems for biological experiments must simultaneously understand experimental protocols, perceive experimental states, and perform fine-grained operations. Existing automated biological experimentation platforms can improve the execution efficiency of standardized workflows, but they typically rely on fixed scripts, dedicated instruments, and structured workflows. This paradigm limits their applicability to open-ended protocols and general-purpose robotic manipulation scenarios. In recent years, foundation models such as LLMs and VLMs have provided new tools for experimental task parsing and state assessment, while VLA models and data augmentation methods have further advanced language-conditioned robotic execution. Therefore, this section reviews related studies from three perspectives: automated biological experimentation, foundation-model-assisted task understanding and verification, and robust robotic manipulation execution.

\subsection{Automated Biological Experiments and Robotic Laboratories}
In recent years, automated experimental systems have gradually evolved from high-throughput instruments with fixed workflows toward robotic laboratories with a certain degree of autonomous planning capability. Szymanski et al. proposed A-Lab, which integrates computational screening, literature knowledge, machine learning, active learning, and robotic experimentation to accelerate the synthesis of inorganic materials~\cite{RN1}. Dai et al. further demonstrated the autonomous operation capability of mobile robots in exploratory synthetic chemistry experiments~\cite{RN9}. Tom et al. reviewed the system architectures, experimental planning strategies, and automated workflows of self-driving laboratories in chemistry and materials science~\cite{RN10}.

In the field of biological laboratory automation, Stephenson et al. summarized physical laboratory automation technologies in synthetic biology~\cite{RN11}. Anhel et al. proposed the Laboratory Automation Protocol format and repository to improve the reusability and standardization of synthetic biology workflows~\cite{RN12}. Jiang et al. introduced ProtoCode, which converts PCR protocols from scientific literature into machine-readable representations~\cite{RN3}. In studies more closely related to autonomous biological experimentation, Singh et al. proposed an AI-driven autonomous enzyme engineering platform that integrates machine learning, LLMs, and biofoundry automation~\cite{RN14}. Pivin developed OSCAR, a modular open-source robotic platform for life science laboratories~\cite{RN15}. Salazar-Villacis et al. proposed the ADePT framework, which evaluates the capabilities of autonomous laboratory robots from the perspectives of adaptability, dexterity, perception, and task complexity~\cite{RN16}. Collectively, these studies indicate that laboratory automation is moving from single-instrument control toward multi-module coordination, autonomous planning, and closed-loop experimentation.

Although the above studies have advanced automated experimental platforms, most existing systems still rely on fixed hardware, predefined scripts, or specific experimental workflows. Such systems are well suited to standardized and high-throughput tasks, but they are difficult to directly adapt to open-ended biological protocols and general-purpose robotic manipulation scenarios. In real wet-lab environments, robotic manipulators must not only understand natural-language protocols, but also recognize the states of experimental labware, determine whether task execution conditions are satisfied, and translate protocol steps into continuous robotic actions. Therefore, existing studies on laboratory automation provide the application background for this work, while their dependence on fixed workflows and dedicated instruments also suggests that a system-level framework integrating protocol understanding, visual verification, and embodied execution is still needed for general-purpose robotic manipulators in biological experiments.

\subsection{LLM/VLM-Assisted Experimental Task Understanding and Verification}
Foundation models provide new technical pathways for experimental task understanding and robotic state verification. The application of LLMs in robotics and laboratory automation has first been reflected in natural-language task parsing and script generation. Vemprala et al. used ChatGPT to investigate how prompting strategies and function libraries can adapt LLMs to different types of robotic tasks~\cite{RN17}. Inagaki et al. demonstrated that GPT-4 can generate operation scripts for an OT-2 liquid-handling robot from goal-oriented instructions in biological experiments~\cite{RN18}. O’Donoghue et al. proposed the BioPlanner framework, which establishes a biological protocol planning benchmark and the BioProt dataset to evaluate the protocol planning capability of LLMs~\cite{RN19}. For machine-readable experimental protocols, Yi et al. proposed the ProtoMed-LLM framework, which focuses on pseudocode extraction and automatic evaluation for biological protocols~\cite{RN20}. In the field of scientific experimental agents, Boiko et al. proposed Coscientist, demonstrating the capability of a GPT-4-driven system in chemical experiment design, planning, and execution~\cite{RN21}. Andres et al. proposed the ChemCrow agent, which augments LLMs with chemistry tools to improve their capabilities in synthesis planning and materials design~\cite{RN22}.

Meanwhile, VLMs have attracted increasing attention in robotic scene understanding, task-state assessment, and execution outcome verification. Driess et al. proposed PaLM-E, which integrates visual observations, state information, and textual instructions as inputs to a multimodal language model for robotic planning, visual question answering, and embodied reasoning tasks~\cite{RN23}. Sermanet et al. proposed RoboVQA and constructed a visual question answering dataset for long-horizon robotic tasks, enabling multimodal reasoning around task planning, success determination, affordance assessment, and future state prediction~\cite{RN24}. Zhang et al. proposed the RoboChemist framework, in which VLMs are used as planners, visual prompt generators, and monitors for experimental robots to perform task decomposition, visual guidance, and execution outcome assessment in chemical experimental tasks~\cite{RN25}. In addition, Retrieval-Augmented Generation (RAG) combines parametric models with external knowledge bases, providing retrievable and updatable information sources for knowledge-intensive tasks and offering a technical foundation for state verification and rule-based judgment in specialized experimental scenarios~\cite{RN26}.

These studies indicate that LLMs are suitable for transforming unstructured experimental texts into structured task descriptions, VLMs are effective for understanding experimental scenes by integrating images and language, and RAG can provide external knowledge support for specialized task assessment. However, existing studies often apply these capabilities separately to protocol generation, robotic script writing, visual question answering, or spatial constraint generation. Few studies have integrated LLM-based protocol parsing, VLM-based scene assessment, knowledge-base retrieval, and robotic state feedback into the execution process of biological experiments. In this work, the LLM is used as the entry point for structured protocol parsing, while VLM-RAG serves as the experimental state verification module. This design enables foundation models to undertake task understanding and state feedback functions within a robotic experimentation system, thereby supporting reliable downstream robotic execution.

\subsection{VLA Models and Data Augmentation for Robust Robotic Execution}
VLA models and imitation learning methods provide a foundation for robotic manipulators to translate language instructions into action execution. Zhao et al. proposed ACT, which alleviates error accumulation in long-horizon imitation learning through action chunking and enables fine-grained manipulation on a low-cost dual-Arm platform~\cite{RN27}. Chi et al. proposed Diffusion Policy, which formulates robotic action generation as a conditional diffusion process and improves the performance of visuomotor policies in multi-task manipulation~\cite{RN28}. Subsequently, Zitkovich et al. proposed RT-2, which transfers Internet-scale vision-language pretraining knowledge to robotic control~\cite{RN29}. For dual-Arm and complex manipulation, Liu et al. proposed RDT-1B, which adopts a diffusion Transformer to construct a foundation model for dual-Arm manipulation~\cite{RN30}. Black et al. proposed $\pi_0$, a general robotic control model built on a pretrained VLM and flow matching~\cite{RN31}. Zheng et al. proposed X-VLA, which uses soft prompts to address data heterogeneity across robotic embodiments~\cite{RN7}, whereas Shukor et al. proposed SmolVLA, emphasizing VLA deployment under low-cost and low-computation conditions~\cite{RN8}.

Dataset construction and data augmentation methods also play an important role in improving the generalization capability of robotic policies. Yu et al. proposed ROSIE, which uses text-to-image generation models to produce semantically imagined experiences and thereby expands the distribution of robotic training data~\cite{RN32}. Chen et al. proposed GenAug, which uses pretrained generative models to construct semantically consistent scene variations, improving the adaptability of robotic policies to novel objects and environments~\cite{RN33}. Chen et al. further introduced a semantically controllable augmentation method that rapidly expands robotic datasets through generative models and validates its contribution to generalization in real robotic manipulation~\cite{RN34}. Kobayashi et al. investigated the role of image data augmentation in bilateral-control-based imitation learning~\cite{RN35}.

Overall, VLA models have substantially improved the ability of robotic manipulators to perform manipulation tasks according to language instructions. However, most existing methods still focus on the direct mapping from visual observations and language instructions to actions, while lacking high-level semantic judgment of experimental preconditions and execution outcomes. In addition, large-scale VLA models typically require high training and inference costs. Although lightweight models are more suitable for low-cost robotic platforms, they remain susceptible to transparent labware, local reflections, illumination changes, and overexposure in wet-lab scenarios. Data augmentation provides an effective approach for mitigating visual distribution shifts, but existing augmentation strategies are mostly designed for general manipulation or sim-to-real transfer, with limited consideration of transparent consumables and complex illumination conditions in biological experiments. Therefore, this study adopts SmolVLA as the embodied execution module and introduces a visual-perturbation-oriented data augmentation strategy to improve the generalization capability and execution stability of the model in fundamental biological experimental operations.

\section{Method}
\subsection{Multi-Agent Collaborative Framework of BioProVLA-Agent}

\begin{figure*}[!t]
  \centering
  \includegraphics[width=0.80\textwidth]{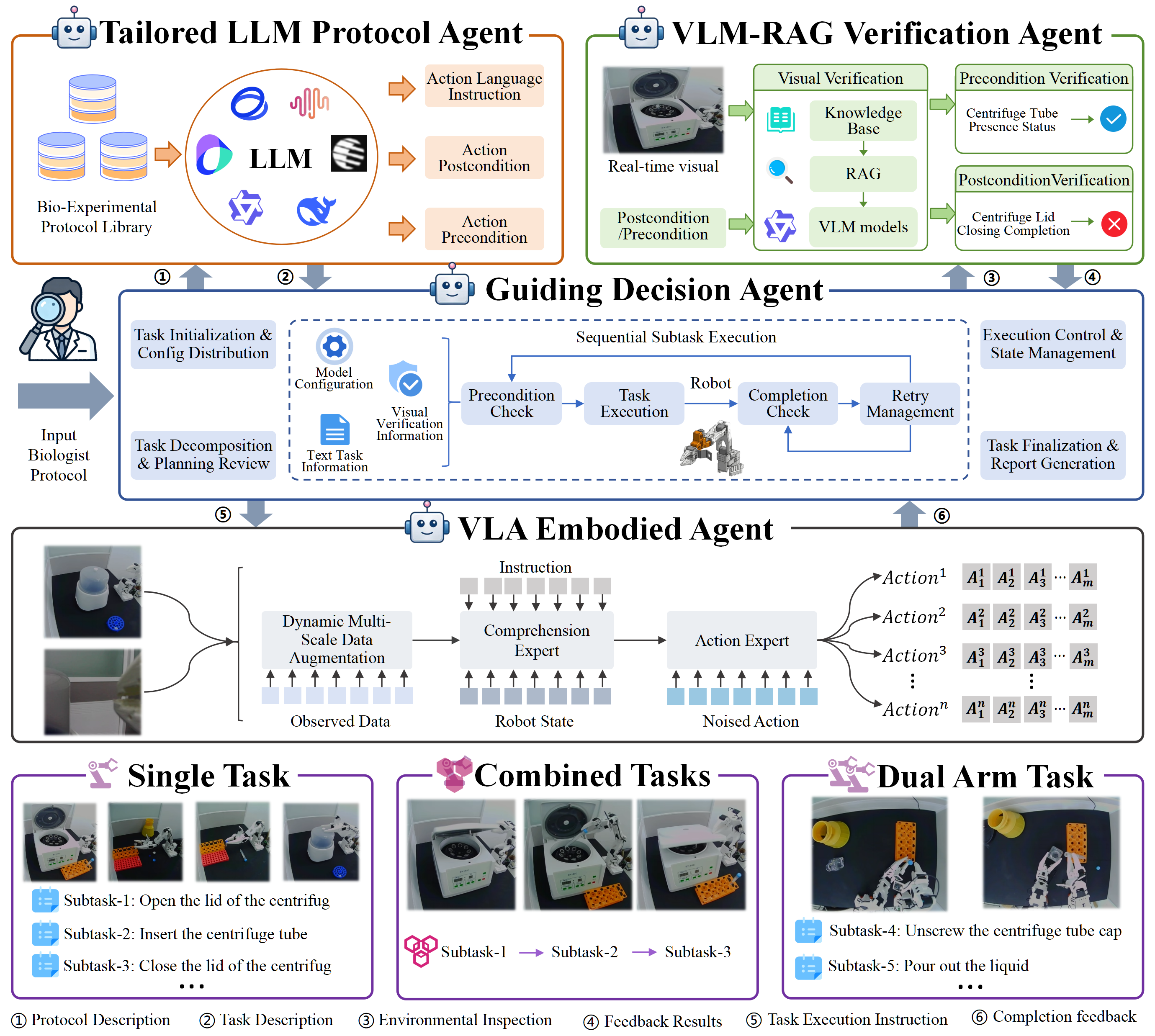}
  \caption{Overall framework of BioProVLA-Agent. The Guiding Decision Agent coordinates protocol parsing, state verification, and embodied execution through the Tailored LLM Protocol Agent, VLM-RAG Verification Agent, and VLA Embodied Agent.}
  \label{fig1}
\end{figure*}

To address the fragmentation among protocol understanding, state verification, and stable execution in robotic manipulators for biological experiments, this study proposes BioProVLA-Agent, a multi-agent collaborative framework, as illustrated in Fig \ref{fig1}. In the overall workflow, the Tailored LLM Protocol Agent first parses the experimental protocol into structured task representations; the VLM-RAG Verification Agent then provides closed-loop constraints through task-state verification; and the VLA Embodied Agent produces the final embodied execution. Throughout this process, the Guiding Decision Agent maintains information flow, state updates, and execution coordination across all stages.

Following this collaborative workflow, BioProVLA-Agent first maps the experimental protocol, model selection, and manipulator configuration into an initial system state shared by all sub-agents. Given an input experimental protocol $P$ a model configuration set $\mathcal{M} = \{M_{\text{LLM}}, M_{\text{VLM}}, M_{\text{VLA}}\}$, and a manipulator configuration $C_R$, the Guiding Decision Agent performs system initialization and resource allocation, which can be formulated as:

\begin{equation}
S_0 = G(P, \mathcal{M}, C_R)
\end{equation}

where $S_0$ denotes the initialized system state, including the model loading status, the initial pose of the robotic manipulator, and the availability status of each sub-agent.

Subsequently, the Guiding Decision Agent sends the experimental protocol to the Tailored LLM Protocol Agent to obtain a structured subtask sequence:

\begin{equation}
\mathcal{T} = \{\tau_i\}_{i=1}^N
\end{equation}

\begin{equation}
\tau_i = (I_i, Pre_i, Post_i, K_i)
\end{equation}

where $I_i$ denotes the natural-language execution instruction of the $i$-th subtask, $Pre_i$ and $Post_i$ denote the task precondition and completion condition, respectively, and $K_i$ denotes the corresponding knowledge-base index. The Guiding Decision Agent checks the subtask order, task completeness, and executability of this sequence to reduce the impact of protocol-parsing errors on subsequent execution.

During the execution stage, the Guiding Decision Agent establishes a closed-loop control process consisting of precondition verification, task execution, and completion verification according to the subtask sequence. For each subtask $\tau_i$, the system first invokes the VLM-RAG Verification Agent to determine whether the precondition is satisfied based on the real-time visual observation $O_t$, the robot state $R_t$, and the knowledge-base index $K_i$:

\begin{equation}
v_i^{pre} = V(O_t, R_t, Pre_i, K_i)
\end{equation}

When $v_i^{pre}=1$, the current experimental state is considered to satisfy the execution requirements. The Guiding Decision Agent then sends the instruction $I_i$ to the VLA Embodied Agent, which generates an action sequence and drives the robotic manipulator to execute the task:

\begin{equation}
A_i = \pi_{VLA}(O_t, R_t, I_i)
\end{equation}

When $v_i^{pre}=0$, the system does not proceed directly to action execution. Instead, the Guiding Decision Agent makes a scheduling decision according to the failure reason returned by the verification module. If an unexecuted subtask in the task sequence can satisfy the required precondition, the execution order is adjusted so that the corresponding preparatory operation is performed first. If the issue cannot be resolved by reordering the sequence, a second verification is triggered to reduce the effect of possible visual misjudgment or state-reading errors. If the precondition remains unsatisfied after rechecking, the current task is suspended and human intervention is requested.

\begin{figure*}[!t]
  \centering
  \includegraphics[width=0.80\textwidth]{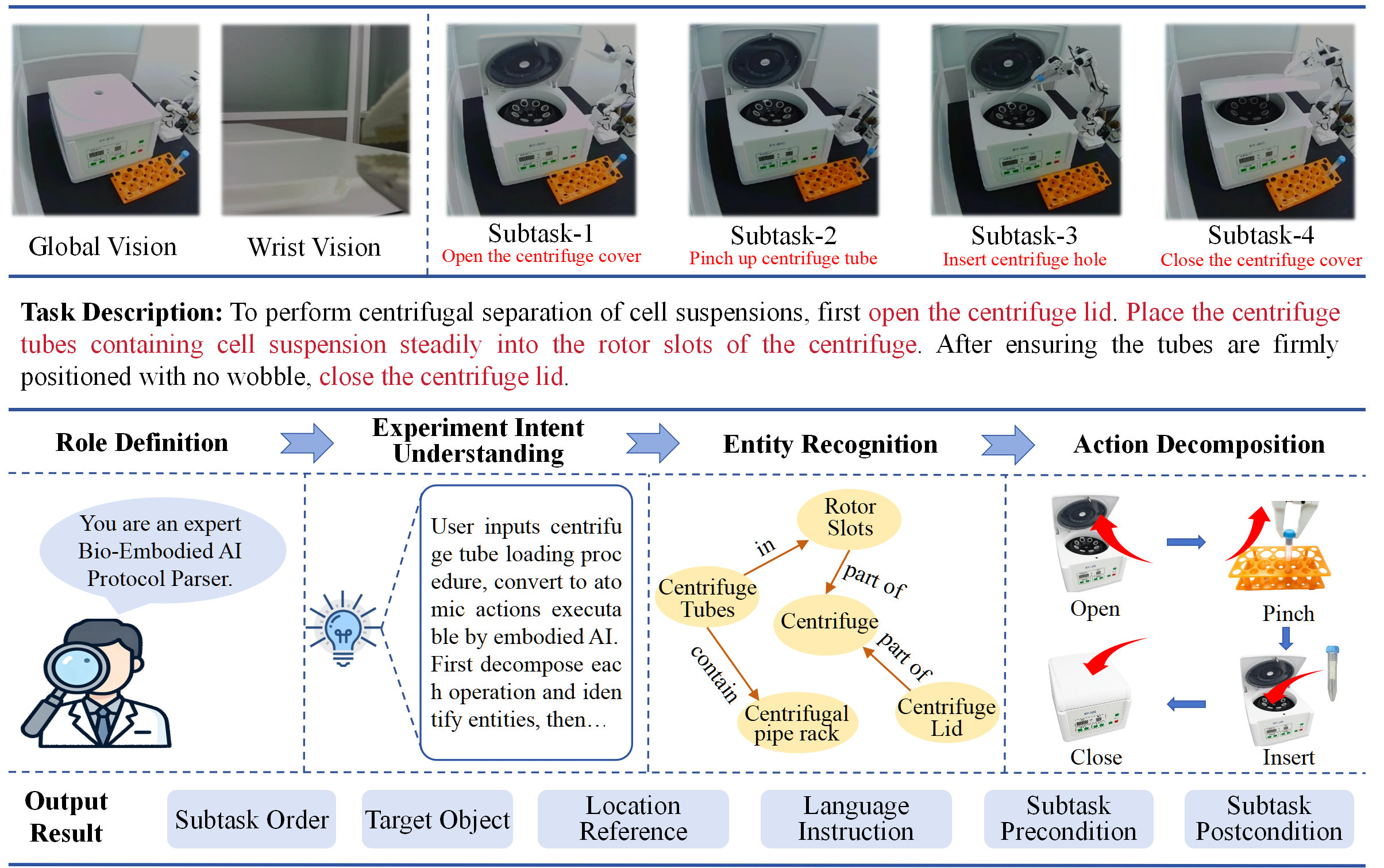}
  \caption{Structured parsing workflow of biological experimental protocols in the Tailored LLM Protocol Agent.}
  \label{fig2}
\end{figure*}

After task execution is completed, the Guiding Decision Agent invokes the VLM-RAG Verification Agent again to verify the completion condition:

\begin{equation}
v_i^{post} = V(O_{t+1}, R_{t+1}, Post_i, K_i)
\end{equation}

When $v_i^{post}=1$, the subtask is marked as completed, and the system proceeds to the next subtask. If the verification fails, the Guiding Decision Agent triggers re-execution, sequence adjustment, or human intervention according to the failure reason and the number of retries. Through this centralized scheduling and closed-loop feedback mechanism, BioVLA-AgentBioProVLA-Agent continuously maintains task states during long-horizon biological experiments, reduces the accumulation of single-step execution errors in subsequent procedures, and improves the reliability and controllability of robotic experimental operations.

\subsection{Tailored LLM Protocol Agent}
The Custom LLM Parse AgentTailored LLM Protocol Agent is designed to transform unstructured biological experimental protocols into executable and verifiable structured subtask sequences. This agent first performs semantic parsing of the protocol text to identify the experimental intent, manipulated objects, spatial relationships, and action order, and then generates an intermediate task representation for downstream verification and execution modules. The overall workflow is shown in Fig \ref{fig2}.

Given an input experimental protocol $P$, the Custom LLM Parse AgentTailored LLM Protocol Agent performs protocol parsing using a prompt-constrained large language model $F_{LLM}$:

\begin{equation}
Y = F_{LLM}(P; \Theta_p)
\end{equation}

where $\Theta_p$ denotes the protocol-parsing prompt set, which includes role definition, experimental intent understanding, entity recognition, action-space constraints, semantic-granularity constraints, and structured output rules. The parsing result $Y$ is not directly used as a robot control command; instead, it serves as an intermediate task representation that can be scheduled by the Guiding Decision Agent.
During parsing, the model is first assigned the role of an embodied-intelligence protocol parsing expert for biological experiments, enabling it to interpret the input text from the perspective of laboratory automation and robotic execution. The model then performs experimental intent parsing on the protocol and extracts the experimental objective corresponding to the current step:

\begin{equation}
g = f_{intent}(P; \Theta_p)
\end{equation}

where $g$ denotes the high-level experimental intent of the protocol. Based on this intent, the model further identifies key entities in the experimental text, including laboratory equipment, consumables, target objects, and location references:

\begin{equation}
E = f_{entity}(P, g; \Theta_p)
\end{equation}

where $E$ denotes the identified entity set. Based on the experimental intent and entity information, the model maps the explicit action predicates in the protocol to a predefined atomic action space:

\begin{equation}
a_i = \phi(v_i, A)
\end{equation}

\begin{figure*}[!t]
  \centering
  \includegraphics[width=0.80\textwidth]{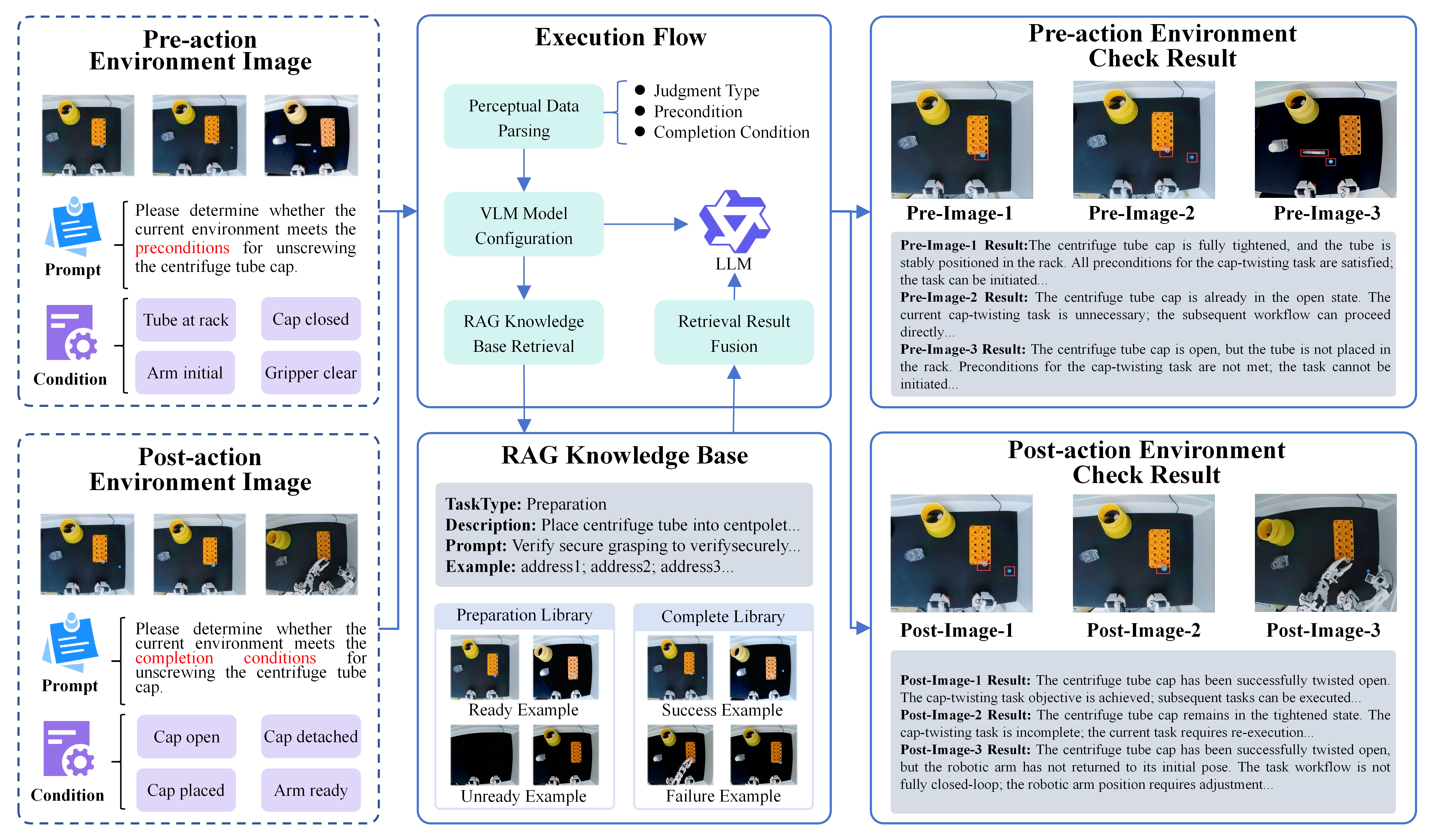}
  \caption{Closed-loop pre- and post-action state verification workflow of the VLM-RAG Verification Agent.}
  \label{fig3}
\end{figure*}

where $v_i$ denotes the $i$-th key action predicate in the protocol, and A is a closed atomic action set that includes basic experimental operations such as lid opening, lid closing, object placement, object removal, grasping, movement, button pressing, and waiting. This constraint reduces the generation of action descriptions that exceed the capability range of the robotic manipulator.

The Custom LLM Parse AgentTailored LLM Protocol Agent outputs the parsing results in a structured JSON format, including the experimental intent, entity recognition results, action decomposition logic, natural-language execution instructions, preconditions, and completion conditions. These results are received by the Guiding Decision Agent and further passed to the VLM-RAG Verification Agent and the VLA Embodied Agent, enabling a stable transformation from experimental protocol semantics to the robotic execution interface.

\subsection{VLM-RAG Verification Agent}
The VLM-RAG Verification Agent performs semantic-level verification of the experimental environment state during robotic execution. Its core objective is to determine whether the current experimental state satisfies the task preconditions before execution and whether the expected completion conditions are achieved after execution. Unlike direct verification based solely on a vision model, this module introduces a retrieval-augmented mechanism that integrates real-time environmental observations with task specifications, verification prompts, and success/failure examples from the biological operation knowledge base, thereby improving the reliability and interpretability of state judgment. The overall workflow is shown in Fig \ref{fig3}.

For the $i$-th task unit, the Guiding Decision Agent sends a verification request to the VLM-RAG Verification Agent:

\begin{equation}
Q_i^j = (C_i^j, K_{i}, j)
\end{equation}

where $j \in \{\mathrm{pre},\mathrm{post}\}$ denotes the verification type. Specifically, $pre$ corresponds to precondition verification before execution, whereas $post$ corresponds to completion-condition verification after execution. $C_i^j$ denotes the condition to be verified, and $K_i$ is the corresponding knowledge-base index. The VLM-RAG Verification Agent first retrieves the knowledge items most relevant to the current task from the knowledge base $K$ based on semantic similarity:

\begin{equation}
\mathcal{B}_i = {\rm TopK}_{b \in \mathcal{K}} {\rm sim}(\phi(Q_i^j), \phi(b))
\end{equation}

where $\phi(\cdot)$ denotes a text or multimodal embedding function, $\operatorname{sim}(\cdot)$ denotes the similarity metric, and $B_i$ contains the task description, verification prompt, and corresponding success and failure reference examples. Through this retrieval process, the model obtains external prior knowledge related to the current operation rather than relying solely on single-frame visual observation for judgment.

Subsequently, the agent integrates the current environmental image $O_t$, robot state $R_t$, condition to be verified $C_i^j$, and retrieved knowledge items $B_i$ to construct the VLM-oriented verification input:

\begin{equation}
X_i^j = \Psi(O_t, R_t, C_i^j, \mathcal{B}_i, j)
\end{equation}

where $\Psi(\cdot)$ denotes the prompt fusion function. This input explicitly specifies whether the model is required to determine task executability before execution or task completion after execution, thereby avoiding semantic confusion between precondition verification and completion-condition verification. The decision output of the VLM-RAG Verification Agent can be formulated as:

\begin{equation}
D_i^j = F_{\rm VLM}(X_i^j) = (v_i^j, r_i^j)
\end{equation}

\begin{figure*}[!t]
  \centering
  \includegraphics[width=0.80\textwidth]{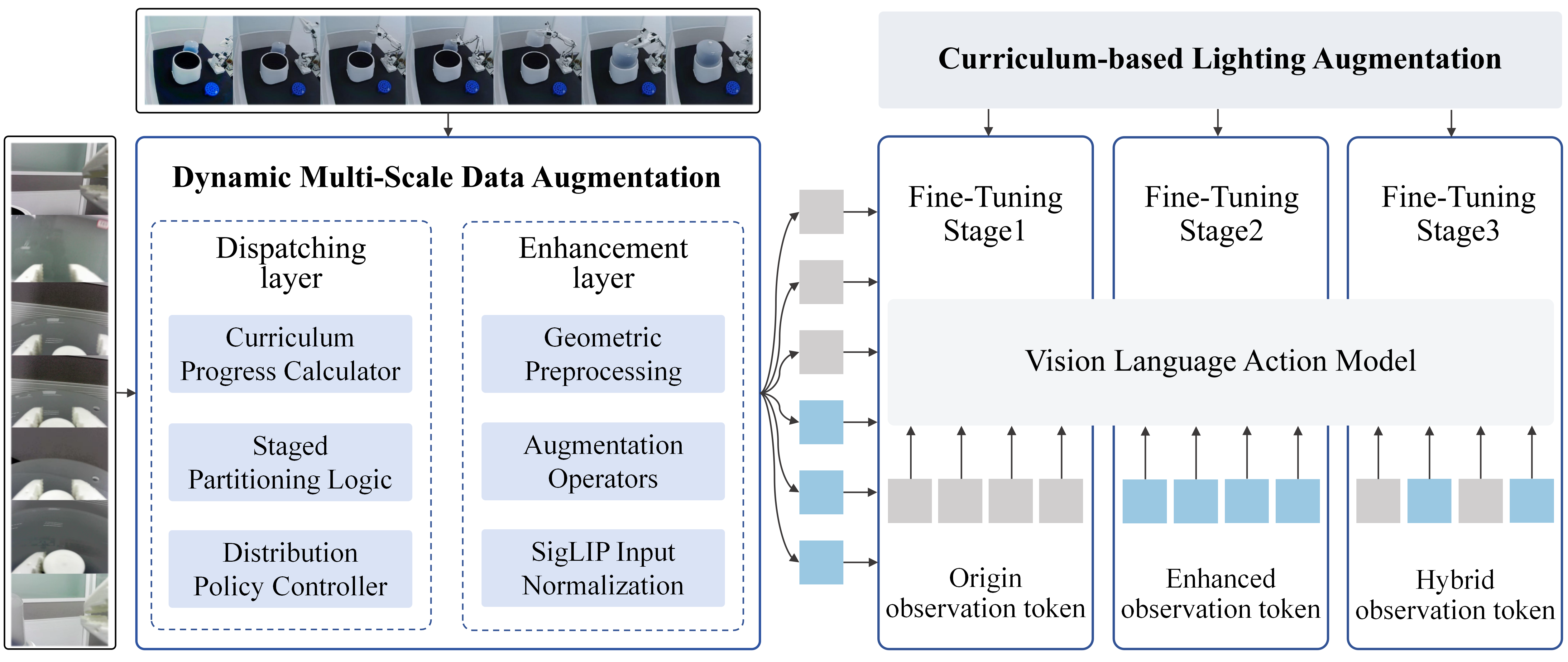}
  \caption{Curriculum-based online data augmentation framework in the VLA Embodied Agent.}
  \label{fig4}
\end{figure*}

where $v_i^j \in \{0,1\}$ is the binary verification result, and $r_i^j$ denotes the corresponding natural-language explanation and failure reason. When $j=pre$ and $v_i^{pre}=1$, the current environment is considered to satisfy the execution requirements, and the Guiding Decision Agent sends the task to the VLA Embodied Agent for execution. If $v_i^{pre}=0$, the agent returns the unsatisfied condition and its reason, after which the Guiding Decision Agent determines whether to adjust the task order, trigger re-verification, or request human intervention. When $j=post$ and $v_i^{post}=1$, the current task is marked as completed. If $v_i^{post}=0$, the system triggers task retry or terminates the process according to the failure reason.

Through this design, the VLM-RAG Verification Agent serves as a visual-semantic monitoring module for task execution in BioVLA-AgentBioProVLA-Agent. It not only determines whether the experimental environment satisfies the operational requirements but also provides interpretable state feedback based on success and failure examples. This enables the robotic system to form a closed-loop verification mechanism consisting of pre-execution feasibility confirmation, post-execution result assessment, and failure-reason feedback during long-horizon biological experiments.

\subsection{VLA Embodied Agent}
The VLA Embodied Agent is responsible for converting verified subtask instructions into action sequences executable by the robotic manipulator, serving as the embodied execution module in BioProVLA-Agent. This module receives natural-language execution instructions from the Guiding Decision Agent and, conditioned on the current visual observation and robot state, invokes the lightweight Vision-Language-Action model SmolVLA to generate continuous control actions. In this way, it performs biological experimental operations such as centrifuge tube placement, labware lid manipulation, waste disposal, and dual-Arm coordination. The training workflow is shown in Fig \ref{fig4}.

Given the current observation $O_t$, robot state $R_t$, and task instruction $I_i$, the action generation process of the VLA Embodied Agent can be formulated as:

\begin{equation}
A_t = \pi_\theta(O_t, R_t, I_i)
\end{equation}

where $\pi_{\theta}$ denotes the SmolVLA policy model, $A_t=\{a_t, a_{t+1}, \ldots, a_{t+H}\}$ is the action sequence output by the model, and $H$ denotes the action prediction horizon. Unlike the high-level agents responsible for protocol parsing and state verification, the VLA Embodied Agent focuses on the low-level mapping from multimodal inputs to robotic actions.

To enhance visual robustness in real wet-lab environments, this study introduces a curriculum-learning-based online data augmentation strategy during SmolVLA fine-tuning. This strategy does not require the offline generation of an additional augmented dataset; instead, it dynamically perturbs input observations during training. Given an original training sample $(O_t,R_t,I_i,A_t^*)$, the online augmented visual observation can be expressed as:

\begin{equation}
\tilde{O}_t = \mathcal{T}_\alpha(O_t)
\end{equation}

where $\mathcal{T}_\alpha$ denotes the augmentation operator composed of lighting perturbations such as brightness adjustment, contrast variation, low illumination, and overexposure, and $\alpha$ represents the perturbation intensity. The training objective is defined as:

\begin{equation}
\mathcal{L}(\theta) = \left\| \pi_\theta(\tilde{O}_t, R_t, I_i) - A_t^* \right\|_2^2
\end{equation}

The above online augmentation strategy expands the visual distribution without increasing data storage cost, allowing the model to encounter more lighting variations and reflection-induced disturbances during training.

\begin{figure}[htbp]
  \centering
  \includegraphics[width=\linewidth]{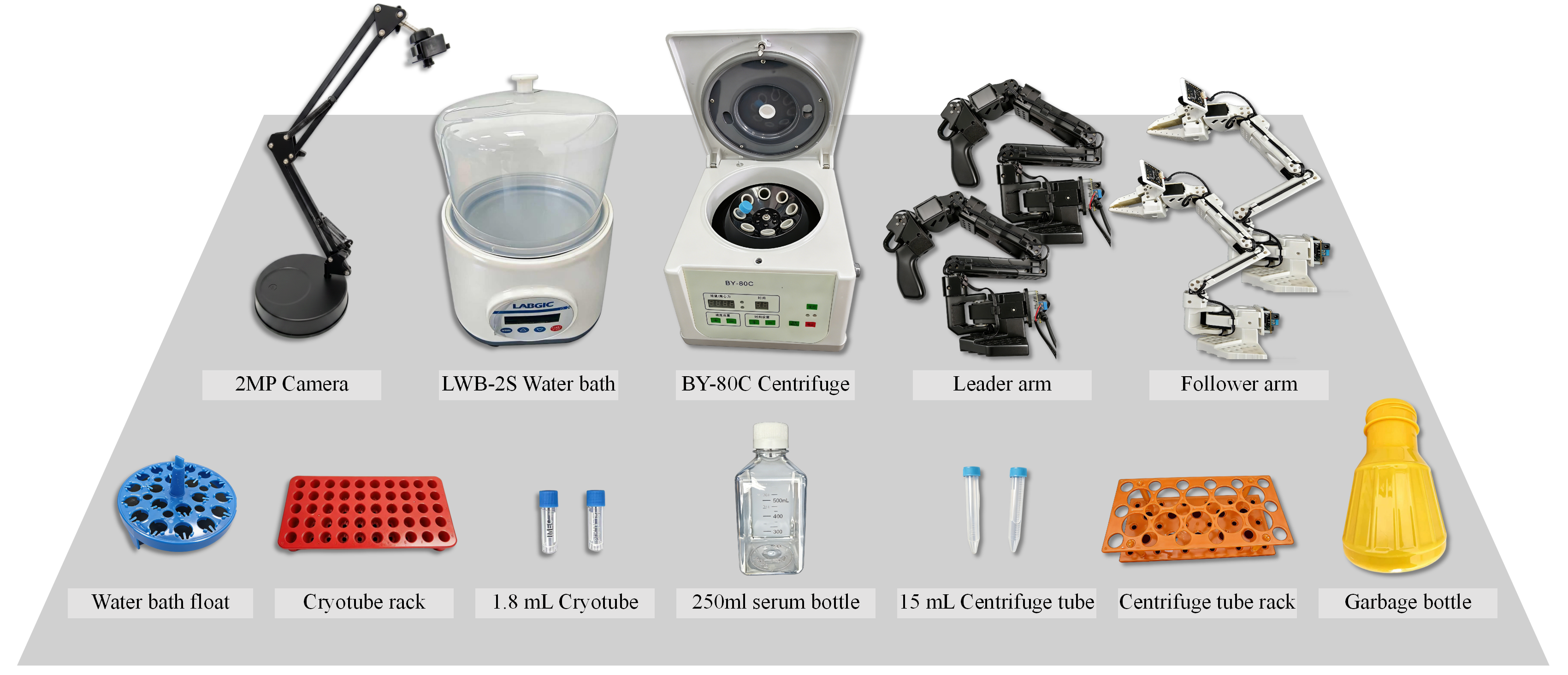}
  \caption{Dataset assets.}
  \label{fig5}
\end{figure}

The augmentation process is further organized as a curriculum-style training strategy~\cite{RN36}. In the early stage, only original observations are used, enabling the model to first learn the action mappings from the original data distribution. In the middle stage, augmented observations under low illumination, strong illumination, and overexposure are gradually introduced to increase the difficulty of visual perturbations. In the final stage, original and augmented observations are jointly used for mixed training, thereby balancing basic execution stability with generalization under complex visual conditions. This process can be formulated as:

\begin{equation}
O_t^{(e)} = 
\begin{cases}
O_t, & e \in \mathcal{E}_1, \\
\mathcal{T}_{\alpha(e)}(O_t), & e \in \mathcal{E}_2, \\
\lambda O_t + (1 - \lambda) \mathcal{T}_{\alpha(e)}(O_t), & e \in \mathcal{E}_3,
\end{cases}
\end{equation}

where e denotes the training stage, and $mathcal{E}_1$, $mathcal{E}_2$, and $mathcal{E}_3$ correspond to the original-data learning stage, augmented-data adaptation stage, and mixed-data consolidation stage, respectively. The perturbation intensity $\alpha(e)$ gradually increases as training proceeds, following the easy-to-hard principle of curriculum learning.

Through this design, the VLA Embodied Agent not only generates robotic actions according to natural-language instructions but also improves the model’s adaptability to lighting variations, reflections from transparent labware, and overexposure disturbances during training.

\section{Experimental Results and Analysis}
\subsection{Dataset}

To reproduce the operational challenges commonly encountered in biological workflows, we constructed a multi-task dataset. All experimental interactions were performed on the So-ARM101 robotic manipulator platform, covering both single-arm and dual-Arm operation modes. The task objects were selected from standardized biological laboratory consumables and equipment. As shown in Fig \ref{fig5}, the main assets include a BY-80C centrifuge, 15 mL centrifuge tubes, 1.8 mL cryotubes, an orange centrifuge tube rack, a red cryotube rack, a water bath, a water-bath float, a 25 mL serum bottle, and a trash bin. The geometric dimensions, assembly tolerances, and physical properties of these objects were kept consistent with those in real wet-lab environments. During data collection, 100 demonstration trajectories were collected for each atomic task via human-in-the-loop teleoperation. All data were stored in the LeRobot format to ensure structural consistency and facilitate subsequent model training and evaluation.

Based on the above assets, we designed 15 atomic tasks around recurrent manipulation actions in biological experiments, as detailed in Table \ref{tab1}. Among them, 12 are single-arm tasks, covering operations such as opening and closing the centrifuge and water-bath lids, precisely placing centrifuge tubes and cryotubes into racks, inserting and removing centrifuge tubes and water-bath floats, and discarding used centrifuge tubes and cryotubes. The remaining three are dual-Arm coordination tasks, including unscrewing and tightening centrifuge tube caps, as well as pouring waste liquid from a tube into a serum bottle. These tasks cover basic manipulations in biological experiments and provide a systematic benchmark for evaluating policy generalization under both single-arm dexterous manipulation and dual-Arm coordination settings.

\begin{table*}[!t]
\caption{Information of atomic tasks.}
\label{tab1}
\centering
\small
\begin{tabularx}{\textwidth}{@{}
>{\raggedright\arraybackslash}p{0.32\textwidth}
>{\raggedright\arraybackslash}p{0.21\textwidth}
>{\raggedright\arraybackslash}p{0.15\textwidth}
>{\raggedright\arraybackslash}X
@{}}
\toprule
\textbf{Task Description} & \textbf{Abbreviation} & \textbf{Task Type} & \textbf{Objects} \\
\midrule
Unlock and open the centrifuge lid
& Open Centrifuge Lid
& Lid manipulation
& BY-80C centrifuge \\

Close the centrifuge lid
& Close Centrifuge Lid
& Lid manipulation
& BY-80C centrifuge \\

Place the centrifuge tube into the centrifuge
& Insert Tube to Centrifuge
& Insertion
& 15 mL centrifuge tube, BY-80C centrifuge, Centrifuge tube rack \\

Remove the centrifuge tube from the centrifuge
& Remove Tube from Centrifuge
& Removal
& 15 mL centrifuge tube, BY-80C centrifuge, Centrifuge tube rack \\

Place the 15 mL centrifuge tube into the orange centrifuge tube rack
& Place Centrifuge Tube to Orange Rack
& Precise placement
& 15 mL centrifuge tube, Centrifuge tube rack \\

Place the 1.8 mL cryotube into the red cryotube rack
& Place Cryotube to Red Rack
& Precise placement
& 1.8 mL cryotube, Cryotube rack \\

Discard the used 15 mL centrifuge tube into the trash can
& Discard Centrifuge Tube
& Disposal
& 15 mL centrifuge tube, Centrifuge tube rack, Garbage bottle \\

Discard the used 1.8 mL cryotube into the trash can
& Discard Cryotube
& Disposal
& 1.8 mL cryotube, Cryotube rack, Garbage bottle \\

Open the water bath lid
& Open Water Bath Lid
& Lid manipulation
& Water bath \\

Close the water bath lid
& Close Water Bath Lid
& Lid manipulation
& Water bath \\

Place the water bath float into the water bath
& Place Float to Water Bath
& Insertion
& Water bath, Water bath float \\

Remove the water bath float from the water bath
& Remove Float from Water Bath
& Removal
& Water bath, Water bath float \\

Unscrew the centrifuge tube cap
& Unscrew Tube Cap
& Twisting operation
& 15 mL centrifuge tube, Centrifuge tube rack \\

Tighten the centrifuge tube cap
& Tighten Tube Cap
& Twisting operation
& 15 mL centrifuge tube, Centrifuge tube rack \\

Pour the waste liquid from the centrifuge tube into the waste liquid bottle
& Pour Waste Liquid
& Clamping and pouring
& 15 mL centrifuge tube, Centrifuge tube rack, 25 mL cerum bottle \\
\bottomrule
\end{tabularx}
\end{table*}

\subsection{Evaluation Metrics}
To systematically evaluate the execution performance of different models in biological robotic manipulation tasks, this study adopts different metrics according to task complexity, including the success rate for single tasks and the completion rate for composite tasks.

For single-task experiments, the Success Rate (SR) is used as the primary evaluation metric. Each atomic task is evaluated through three independent repeated experiments, and the mean success rate and standard error are calculated from the three success-rate values. Let $N_r^+$ denote the number of successful trials in the $r$-th repeated experiment, where the superscript $+$ indicates successfully completed trials, and let $N_r$ denote the total number of trials in that experiment. The success rate of the $r$-th repeated experiment is defined as:

\begin{equation}
SR_r = \frac{N_r^+}{N_r} \times 100\%, \quad r = 1,2,3
\end{equation}

Based on the three repeated experiments, the mean success rate of the model on the corresponding task is calculated as:

\begin{equation}
\overline{SR} = \frac{1}{3} \sum_{r=1}^3 SR_r
\end{equation}

To characterize the statistical variation among repeated experiments, the Standard Error (SE) is further computed. The sample standard deviation of the three success-rate values is first calculated as:

\begin{equation}
S_{SR} = \sqrt{\frac{1}{3-1} \sum_{r=1}^3 (SR_r - \overline{SR})^2}
\end{equation}

where $S_{SR}$ denotes the sample standard deviation of the success rate. Since the number of repeated experiments is $n=3$, the standard error is defined as:

\begin{equation}
SE_{SR} = \frac{S_{SR}}{\sqrt{3}}
\end{equation}

The single-task experimental result is reported as:

\begin{equation}
\overline{SR} \pm SE_{SR}
\end{equation}

where $\overline{SR}$ is retained to two decimal places and expressed as a percentage, and $SE_{SR}$ is also retained to two decimal places. This metric is mainly used to evaluate the execution reliability of the model in atomic operations such as opening, closing, insertion, removal, precise placement, and disposal.

For composite-task experiments, the Completion Rate (CR) is used as the evaluation metric. Each composite task consists of $M$ atomic steps, where $M=2$ or $M=3$. In the $t$-th composite-task trial, the execution result of the $k$-th atomic step is denoted as $x_{t,k}$. If the step is successfully completed, $x_{t,k}=1$; otherwise, $x_{t,k}=0$. Therefore, the completion rate of the t-th trial is defined as:

\begin{equation}
CR_t = \frac{\sum_{k=1}^M x_{t,k}}{M} \times 100\%
\end{equation}

For each type of composite task, 20 repeated trials are conducted, and the arithmetic mean of the 20 single-trial completion rates is used as the average composite-task completion rate of the model:

\begin{equation}
\overline{CR} = \frac{1}{20} \sum_{t=1}^{20} CR_t
\end{equation}

This calculation is equivalent to measuring the proportion of successfully completed atomic steps among all atomic steps across the 20 trials:

\begin{equation}
\overline{CR} = \frac{\sum_{t=1}^{20} \sum_{k=1}^M x_{t,k}}{20M} \times 100\%
\end{equation}

Unlike the success rate used for single tasks, CR does not require all steps in a composite task to be completed successfully. Instead, it quantifies the degree of task completion at the step level. Therefore, this metric provides a more fine-grained assessment of the model’s task-transition capability, state-maintenance ability, and resistance to error accumulation in multi-step biological manipulation tasks, making it suitable for evaluating the overall execution performance of composite tasks and long-horizon robotic experiments.

\subsection{Performance Evaluation of LLM-Based Protocol Parsing}

\begin{figure*}[!t]
  \centering

  \begin{subfigure}[t]{0.51\textwidth}
    \centering
    \includegraphics[width=\linewidth]{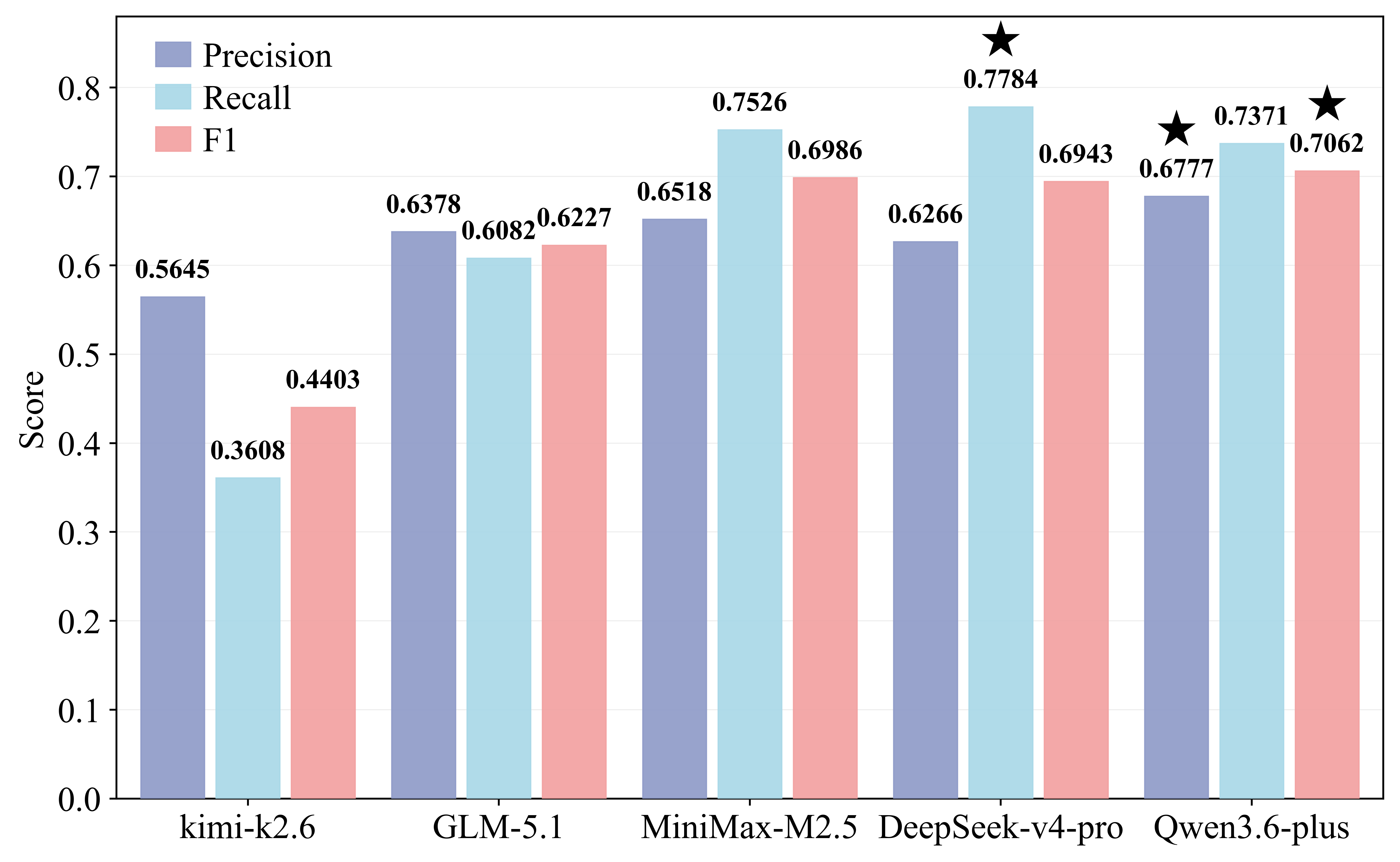}
    \caption{Executable Action Parsing}
    \label{fig8a}
  \end{subfigure}
  \hfill
  \begin{subfigure}[t]{0.495\textwidth}
    \centering
    \includegraphics[width=\linewidth]{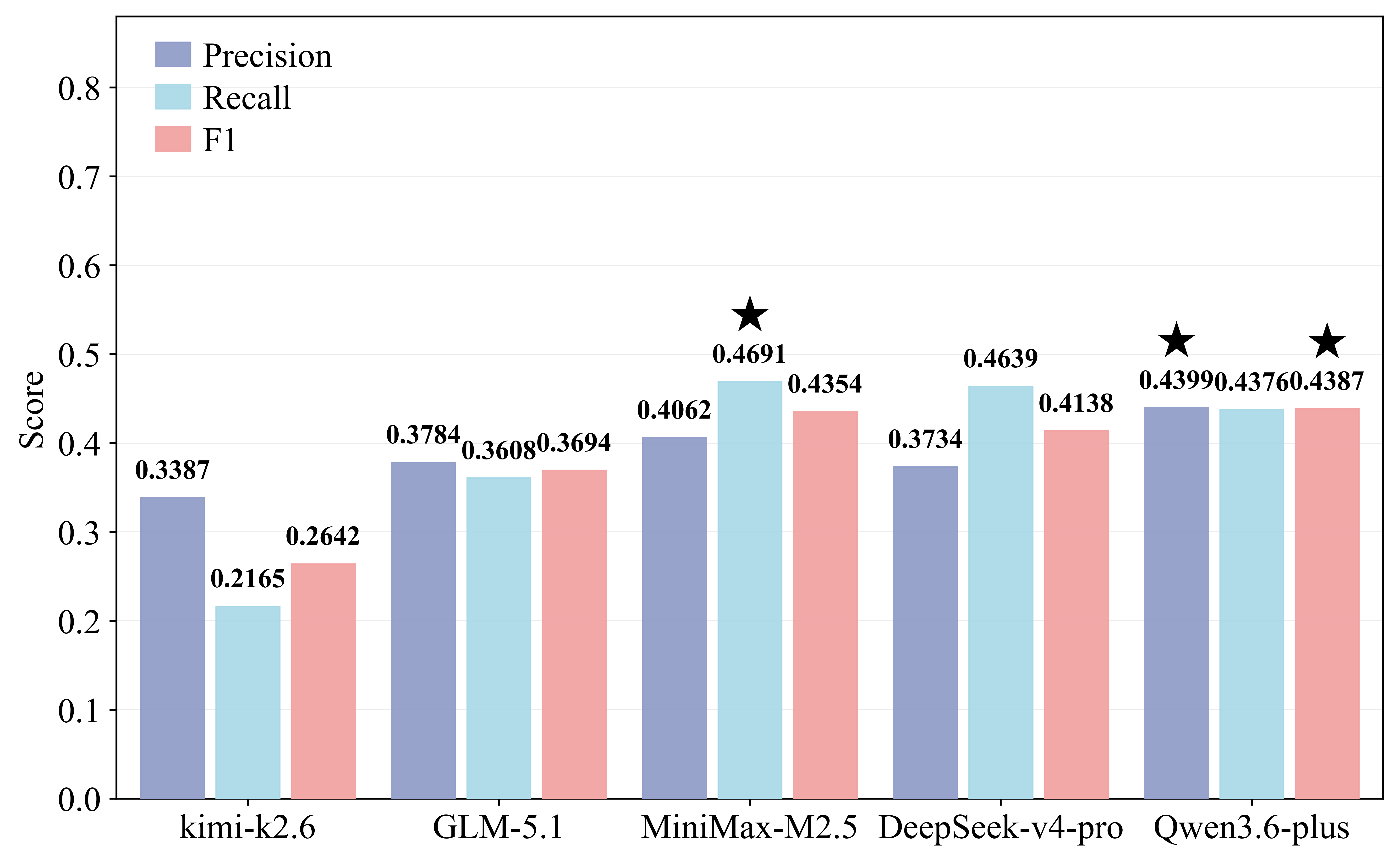}
    \caption{Precondition Generation}
    \label{fig8b}
  \end{subfigure}
  \hfill
  \begin{subfigure}[t]{0.495\textwidth}
    \centering
    \includegraphics[width=\linewidth]{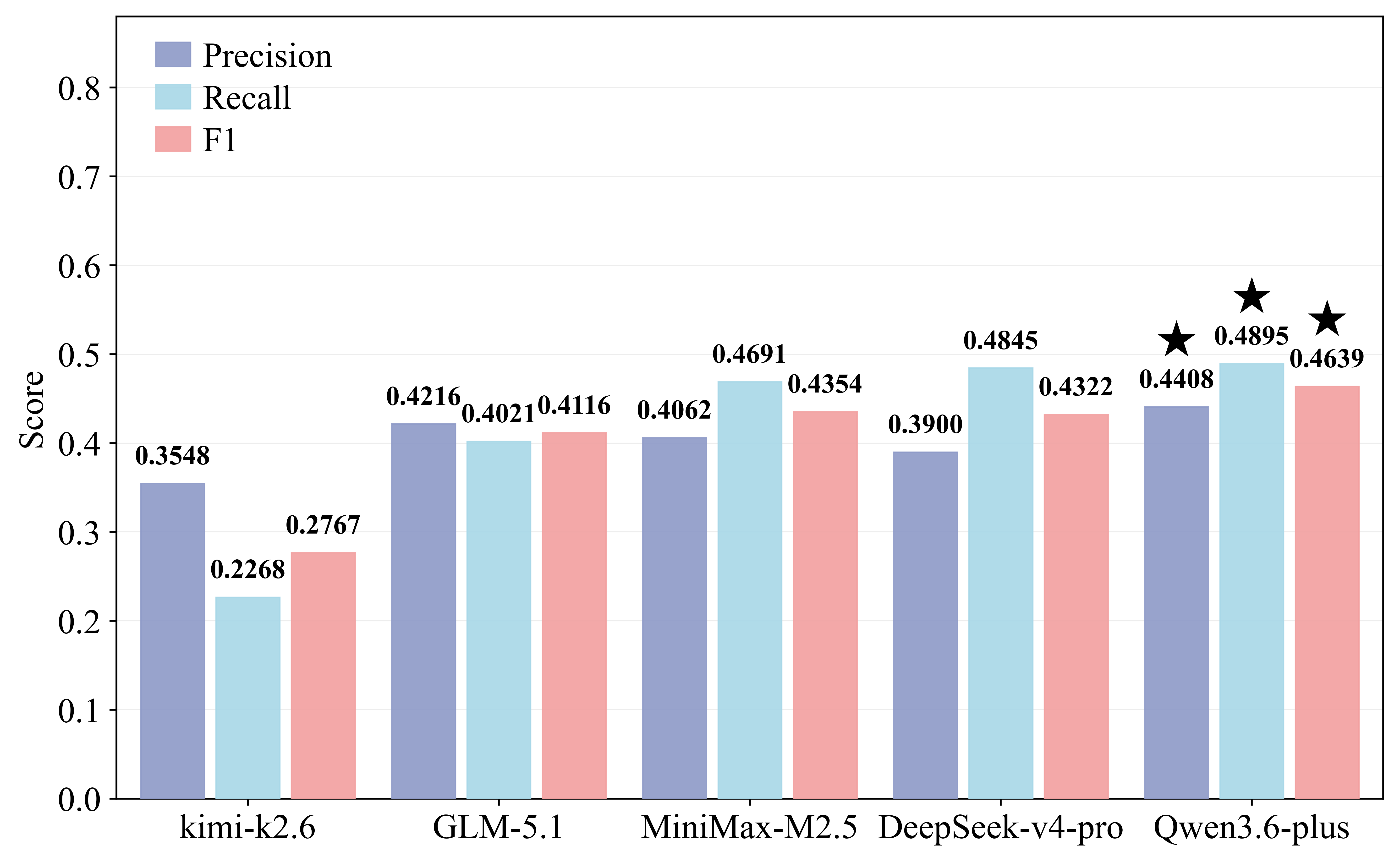}
    \caption{Postcondition Generation}
    \label{fig8c}
  \end{subfigure}

  \caption{Performance comparison of different LLM on executable action parsing, precondition generation, and postcondition generation tasks.}
  \label{fig8}
\end{figure*}

To select a large language model capable of stably transforming unstructured biological protocols into executable and verifiable task representations, a quantitative evaluation was conducted for the Tailored LLM Protocol Agent. A manually annotated protocol-parsing dataset was used as the reference set, in which each protocol was labeled with executable atomic action sequences, preconditions, and postconditions. Five representative LLMs, including Kimi-K2.6, GLM-5.1, MiniMax-M2.5, DeepSeek-V4-Pro, and Qwen3.6-Plus, were evaluated under the same prompt template, predefined action space, and structured JSON output constraints. The generated results were compared with the reference annotations, and precision, recall, and F1 score were calculated separately for executable action parsing, precondition generation, and postcondition generation.

As shown in Fig \ref{fig8}, Qwen3.6-Plus achieved the best overall performance in executable action parsing, with a precision of 0.6777, recall of 0.7371, and F1 score of 0.7062. MiniMax-M2.5 obtained a comparable F1 score of 0.6986, with higher recall of 0.7526 but slightly lower precision of 0.6518. DeepSeek-V4-Pro achieved the highest recall of 0.7784, but its precision was 0.6266, indicating that it tended to generate more candidate actions. GLM-5.1 produced a relatively balanced result with an F1 score of 0.6227, whereas Kimi-K2.6 showed the lowest recall of 0.3608 and an F1 score of 0.4403. For precondition generation, Qwen3.6-Plus again achieved the highest F1 score of 0.4387, followed by MiniMax-M2.5 with 0.4354 and DeepSeek-V4-Pro with 0.4138. GLM-5.1 and Kimi-K2.6 obtained lower F1 scores of 0.3694 and 0.2642, respectively. In postcondition generation, Qwen3.6-Plus remained the best-performing model, reaching a precision of 0.4408, recall of 0.4895, and F1 score of 0.4639. MiniMax-M2.5 and DeepSeek-V4-Pro achieved similar F1 scores of 0.4354 and 0.4322, while GLM-5.1 and Kimi-K2.6 obtained 0.4116 and 0.2767, respectively.

These results indicate that executable action parsing is consistently less difficult than precondition and postcondition generation across the evaluated models. This difference reflects the unequal complexity of the three parsing outputs: atomic actions are usually anchored by explicit operation verbs and physical entities in the protocol, whereas preconditions and postconditions require the model to formulate visual-state constraints around each action, such as object visibility, lid state, and final object placement. Among the compared models, Qwen3.6-Plus provided the most balanced performance across all three components and was therefore selected as the parsing backbone of the Tailored LLM Protocol Agent.

\subsection{Evaluation of the VLM-RAG Verification Agent}

After protocol parsing, reliable robotic execution depends on whether the system can correctly judge the visual state of the experimental environment before and after each subtask. Therefore, this section examines the state-verification capability of the VLM-RAG Verification Agent, which serves as the visual-semantic checking module in the closed-loop workflow. The experiment was constructed using task-state images corresponding to pre-execution and post-execution situations. Specifically, 39 precondition verification samples and 36 postcondition verification samples were included, resulting in 75 verification cases in total. Each model was required to determine whether the given visual observation satisfied the corresponding textual condition. The prediction was then compared with the manually annotated label, and the verification accuracy was calculated for precondition verification, postcondition verification, and the overall task set.

\begin{table*}[!t]
\centering
\caption{Performance comparison of VLM-based precondition and postcondition verification.}
\label{tab7}
\small
\renewcommand{\arraystretch}{1.20}
\begin{tabular*}{\textwidth}{@{\extracolsep{\fill}}lccc@{}}
\toprule
\textbf{Model} 
& \makecell{\textbf{Precondition Generation}} 
& \makecell{\textbf{Postcondition Generation}} 
& \textbf{Overall} \\
\midrule
Qwen3-VL-thinking 
& 30.77\% (12/39) 
& 47.22\% (17/36) 
& 38.67\% (29/75) \\

Kimi-K2.6 
& 66.67\% (26/39) 
& 66.67\% (24/36) 
& 66.67\% (50/75) \\

GLM-5V-Turbo 
& 64.10\% (25/39) 
& 72.22\% (26/36) 
& 68.00\% (51/75) \\

Doubao-Seed-2.0-Pro 
& 71.79\% (28/39) 
& 75.00\% (27/36)
& 73.33\% (55/75) \\

RAG-Doubao-Seed-2.0-Pro 
& \textbf{84.62\% (33/39)} 
& \textbf{80.56\% (29/36)} 
& \textbf{82.67\% (62/75)} \\
\bottomrule
\end{tabular*}
\end{table*}

As shown in Table \ref{tab7}, the retrieval-augmented model achieved the best overall performance. RAG-Doubao-Seed-2.0-Pro correctly verified 33 of 39 precondition cases, reaching 84.62\%, and 29 of 36 postcondition cases, reaching 80.56\%. Its overall accuracy was 82.67\% with 62 correct judgments among 75 cases. Compared with the original Doubao-Seed-2.0-Pro, which achieved 71.79\% in precondition verification, 75.00\% in postcondition verification, and 73.33\% overall accuracy, the RAG-enhanced version improved the overall result by 9.34 percentage points. This improvement indicates that retrieved task knowledge and reference examples provide useful contextual constraints for visual state judgment, especially in determining whether the current scene satisfies the execution requirements before action execution. Among the non-RAG models, Doubao-Seed-2.0-Pro achieved the strongest overall result of 73.33\%, followed by GLM-5V-Turbo with 68.00\% and Kimi-K2.6 with 66.67\%. Qwen3-VL-Thinking showed the weakest performance, with an overall accuracy of 38.67\%.

The comparison between precondition and postcondition verification further reveals differences in the difficulty of the two verification stages. Most models performed better on postcondition verification than on precondition verification. For example, GLM-5V-Turbo increased from 64.10\% on preconditions to 72.22\% on postconditions, and Qwen3-VL-thinking increased from 30.77\% to 47.22\%. This may be because postconditions often correspond to more explicit visual outcomes, such as a lid being closed, an object being placed in a target position, or a container state being changed after manipulation. In contrast, precondition verification requires the model to judge whether the current scene is ready for action, which often involves more subtle spatial and state cues. The performance gain of RAG-Doubao-Seed-2.0-Pro suggests that retrieval-augmented visual reasoning is beneficial for reducing such ambiguity and provides a more reliable verification interface for the subsequent VLA Embodied Agent.

\subsection{Single-Arm Task Experiments}

\begin{figure*}[!t]
  \centering
  \includegraphics[width=\textwidth]{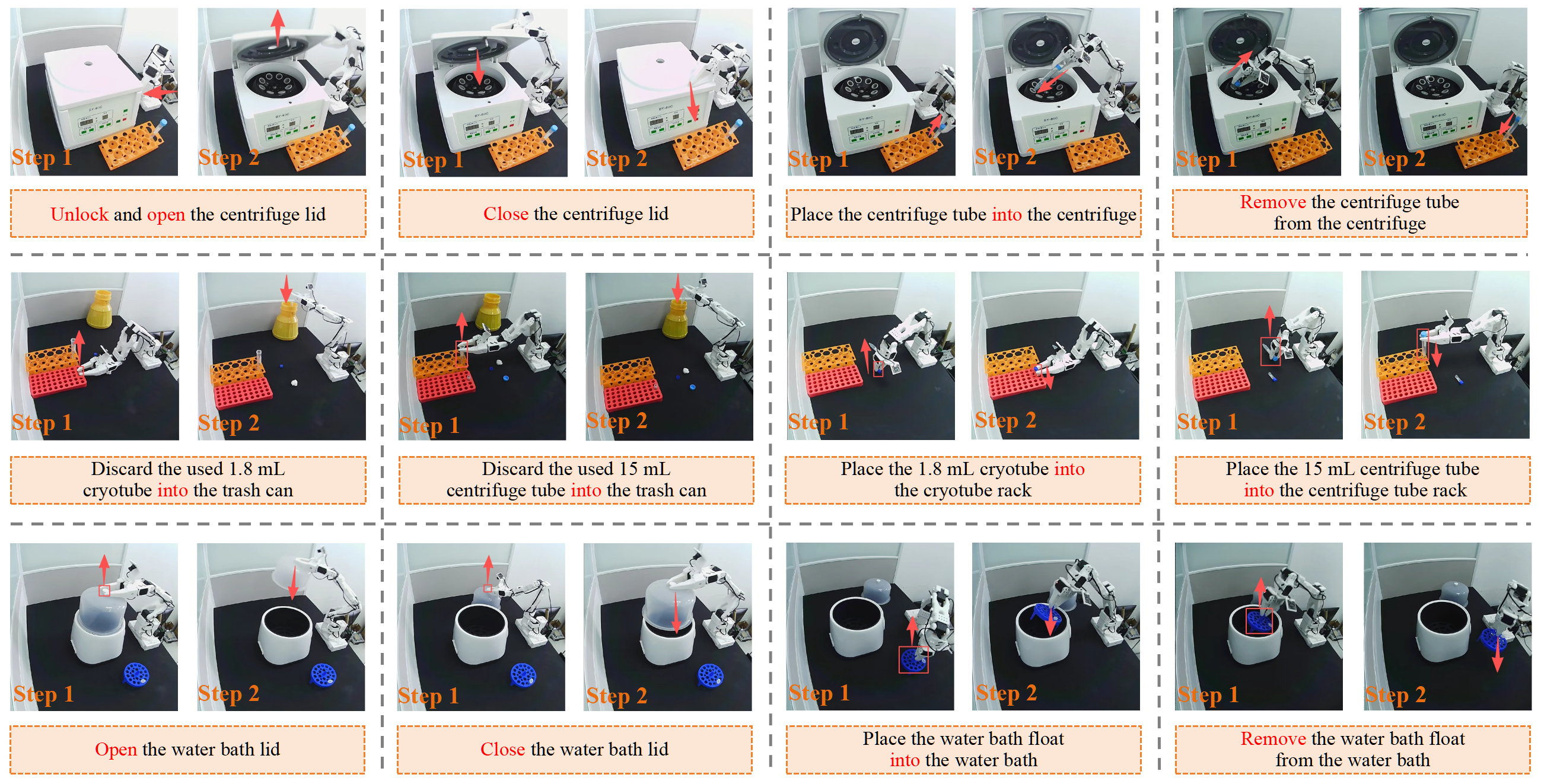}
  \caption{Examples of single-task execution processes.}
  \label{fig6}
\end{figure*}

To systematically evaluate the performance of the proposed method at the basic manipulation level, single-task experiments were conducted on 12 atomic tasks in biological experiments. All tasks were trained and evaluated on the same dataset, with Success Rate (SR) used as the evaluation metric. The compared methods include ACT, X-VLA, and SmolVLA. Each task was executed independently to eliminate inter-task interference, thereby objectively reflecting the capability differences of different models across operation types, such as opening/closing, insertion, precise placement, and disposal. The experimental results are shown in Table \ref{tab2}, and visual examples of the single-task execution process are presented in Fig \ref{fig6}.

Overall, introducing data augmentation improved the performance of SmolVLA across different single-task categories. The success rates of Insert Tube and Remove Tube increased from 43.33\% to 48.33\% and from 51.67\% to 53.33\%, respectively. The success rates of Place Centrifuge Tube and Place Cryotube increased from 43.33\% to 53.33\% and from 40.00\% to 65.00\%, respectively. Performance improvements were also observed in opening/closing tasks involving the centrifuge and water bath lids, as well as in disposal tasks. These results indicate that data augmentation improves model stability and generalization across different operation types. This improvement can be attributed to the lighting variations and local reflection disturbances present in real experimental environments, where the original data distribution cannot cover all observation conditions. By simulating diverse visual scenarios, data augmentation effectively expands the visual distribution during training and improves the model’s adaptability to different lighting conditions.

Considering the observation distribution shift caused by lighting variations in real experimental environments, the effect of data augmentation is particularly evident in precise placement and disposal tasks. AugSmolVLA improved the success rates of Discard Centrifuge Tube and Discard Cryotube from 88.33\% to 90.00\% and from 48.33\% to 55.00\%, respectively. Similarly, the success rates of Place Centrifuge Tube to Orange Rack and Place Cryotube to Red Rack increased from 43.33\% to 53.33\% and from 40.00\% to 65.00\%, respectively. These four tasks involve precise localization and manipulation of transparent objects, which are susceptible to lighting changes, reflections, and background interference. The results indicate that the data augmentation strategy improves localization accuracy and manipulation stability in visually challenging tasks, leading to more robust performance when handling transparent objects.

\begin{table*}[!t]
\caption{Comparison of single-task experimental results.}
\label{tab2}
\centering
\small
\renewcommand{\arraystretch}{1.25}
\begin{tabular*}{\textwidth}{@{\extracolsep{\fill}}lcccc@{}}
\toprule
\textbf{Model} 
& \makecell{\textbf{Open Centrifuge}\\\textbf{Lid (SR)}} 
& \makecell{\textbf{Close Centrifuge}\\\textbf{Lid (SR)}} 
& \makecell{\textbf{Insert Tube to}\\\textbf{Centrifuge (SR)}} 
& \makecell{\textbf{Remove Tube from}\\\textbf{Centrifuge (SR)}} \\
\midrule
Act        & \(43.33\% \pm 4.41\) & \(91.67\% \pm 1.67\) & \(16.67\% \pm 4.41\) & \(23.33\% \pm 3.33\) \\
X-vla      & \(31.67\% \pm 1.67\) & \(43.33\% \pm 3.33\) & \(33.33\% \pm 4.41\) & \(26.67\% \pm 3.33\) \\
Smolvla    & \(53.33\% \pm 6.01\) & \(\mathbf{100.00\% \pm 0.00}\) & \(43.33\% \pm 1.67\) & \(51.67\% \pm 1.67\) \\
AugSmolVLA & \(\mathbf{56.67\% \pm 1.67}\) & \(\mathbf{100.00\% \pm 0.00}\) & \(\mathbf{48.33\% \pm 1.67}\) & \(\mathbf{53.33\% \pm 1.67}\) \\
\midrule
\textbf{Model} 
& \makecell{\textbf{Discard}\\\textbf{Centrifuge}\\\textbf{Tube (SR)}} 
& \makecell{\textbf{Discard}\\\textbf{Cryotube (SR)}} 
& \makecell{\textbf{Place Centrifuge Tube to}\\\textbf{Orange Rack (SR)}} 
& \makecell{\textbf{Place Cryotube to Red}\\\textbf{Rack (SR)}} \\
\midrule
Act        & \(81.67\% \pm 1.67\) & \(35.00\% \pm 2.89\) & \(21.67\% \pm 4.41\) & \(16.67\% \pm 3.33\) \\
X-vla      & \(63.33\% \pm 6.67\) & \(40.00\% \pm 5.00\) & \(30.00\% \pm 5.00\) & \(26.67\% \pm 6.01\) \\
Smolvla    & \(88.33\% \pm 1.67\) & \(48.33\% \pm 4.41\) & \(43.33\% \pm 3.33\) & \(40.00\% \pm 2.89\) \\
AugSmolVLA & \(\mathbf{90.00\% \pm 0.00}\) & \(\mathbf{55.00\% \pm 2.89}\) & \(\mathbf{53.33\% \pm 1.67}\) & \(\mathbf{65.00\% \pm 2.00}\) \\
\midrule
\textbf{Model} 
& \makecell{\textbf{Open Water Bath}\\\textbf{Lid (SR)}} 
& \makecell{\textbf{Close Water}\\\textbf{Bath Lid (SR)}} 
& \makecell{\textbf{Place Float to Water}\\\textbf{Bath (SR)}} 
& \makecell{\textbf{Remove Float from}\\\textbf{Water Bath (SR)}} \\
\midrule
Act        & \(53.33\% \pm 4.41\) & \(61.67\% \pm 7.26\) & \(41.67\% \pm 3.33\) & \(36.67\% \pm 4.41\) \\
X-vla      & \(61.67\% \pm 7.26\) & \(45.00\% \pm 8.66\) & \(26.67\% \pm 3.33\) & \(16.67\% \pm 1.67\) \\
Smolvla    & \(\mathbf{78.33\% \pm 4.41}\) & \(73.33\% \pm 1.67\) & \(43.33\% \pm 4.41\) & \(46.67\% \pm 4.41\) \\
AugSmolVLA & \(\mathbf{78.33\% \pm 1.67}\) & \(\mathbf{75.00\% \pm 2.89}\) & \(\mathbf{46.67\% \pm 4.41}\) & \(\mathbf{50.00\% \pm 4.00}\) \\
\bottomrule
\end{tabular*}
\end{table*}

\subsection{Composite-Task Experiments}
Based on the single-task experiments that evaluated the basic manipulation capability of each model, we further designed composite-task experiments to assess the overall execution performance of the proposed system in multi-step continuous manipulation scenarios. Unlike single-task experiments, which focus only on the success rate of individual atomic operations, composite tasks are closer to real biological experimental workflows and impose higher requirements on task transition, state maintenance, and continuous execution stability. Therefore, multi-step operation procedures were constructed according to the composite-task sequences defined in Table \ref{tab3}, and different methods were evaluated under the same experimental settings. The results are shown in Table \ref{tab4}.

\begin{table*}[!t]
\caption{Definitions of composite-task sequences.}
\label{tab3}
\centering
\small
\renewcommand{\arraystretch}{1.25}
\begin{tabularx}{\textwidth}{@{}
>{\raggedright\arraybackslash}p{0.20\textwidth}
>{\raggedright\arraybackslash}X
>{\raggedright\arraybackslash}p{0.15\textwidth}
@{}}
\toprule
\textbf{Composite task} & \textbf{Subtask Sequence} & \textbf{Task Category} \\
\midrule

Loading centrifuge tube
& \textbf{A:} Open Centrifuge Lid \textbf{B:} Insert Tube to Centrifuge \textbf{C:} Close Centrifuge Lid
& Loading \\

Unload centrifuge tube
& \textbf{A:} Open Centrifuge Lid \textbf{B:} Remove Tube from Centrifuge \textbf{C:} Close Centrifuge Lid
& Unloading \\

Tidy up the desktop
& \textbf{A:} Place Centrifuge Tube to Orange Rack \textbf{B:} Place Cryotube to Red Rack
& Sorting \\

Clean up waste materials
& \textbf{A:} Discard Centrifuge Tube \textbf{B:} Discard Cryotube
& Disposal \\

Loading float
& \textbf{A:} Open Water Bath Lid \textbf{B:} Place Float to Water Bath \textbf{C:} Close Water Bath Lid
& Loading \\

Unload the float
& \textbf{A:} Open Water Bath Lid \textbf{B:} Remove Float from Water Bath \textbf{C:} Close Water Bath Lid
& Unloading \\

\bottomrule
\end{tabularx}
\end{table*}

Unload the float	A: Open Water Bath Lid B: Remove Float from Water Bath C: Close Water Bath Lid	Unloading
Based on the above composite-task settings, model performance in multi-step continuous execution can be evaluated using the Completion Rate (CR). The experimental results show clear differences among the compared methods in composite tasks, with AugSmolVLA achieving the best performance in all six composite-task settings. Specifically, AugSmolVLA obtained completion rates of 56.67\%, 55.00\%, 72.50\%, 75.00\%, 56.67\%, and 58.33\% in Loading centrifuge tube, Unload centrifuge tube, Tidy up the desktop, Clean up waste materials, Loading float, and Unload the float, respectively. Compared with ACT and X-VLA, AugSmolVLA showed more stable performance across loading, unloading, sorting, and disposal tasks. For example, in the Loading centrifuge tube task, AugSmolVLA reached 56.67\%, clearly outperforming ACT and X-VLA, which achieved 28.33\% and 35.00\%, respectively. In the Tidy up the desktop task, AugSmolVLA achieved 72.50\%, exceeding SmolVLA by 10.00 percentage points and showing stronger capability in sequential precise placement. In the Clean-up waste materials task, AugSmolVLA further achieved the highest completion rate of 75.00\%, indicating improved robustness in continuous grasping, transfer, and disposal operations.

\begin{table*}[!t]
\caption{Comparison of composite-task experimental results.}
\label{tab4}
\centering
\small
\renewcommand{\arraystretch}{1.25}
\begin{tabular*}{\textwidth}{@{\extracolsep{\fill}}lcccccc@{}}
\toprule
\textbf{Model}
& \makecell{\textbf{Loading}\\\textbf{centrifuge}\\\textbf{tube (CR)}} 
& \makecell{\textbf{Unload}\\\textbf{centrifuge}\\\textbf{tube (CR)}} 
& \makecell{\textbf{Tidy up the}\\\textbf{desktop (CR)}} 
& \makecell{\textbf{Clean up waste}\\\textbf{materials (CR)}} 
& \makecell{\textbf{Loading}\\\textbf{float (CR)}} 
& \makecell{\textbf{Unload the}\\\textbf{float (CR)}} \\
\midrule
Act        & \(28.33\%\) & \(33.33\%\) & \(35.00\%\) & \(60.00\%\) & \(26.67\%\) & \(31.67\%\) \\
X-vla      & \(35.00\%\) & \(40.00\%\) & \(47.50\%\) & \(40.00\%\) & \(20.00\%\) & \(38.33\%\) \\
Smolvla    & \(55.00\%\) & \(53.33\%\) & \(62.50\%\) & \(62.50\%\) & \(50.00\%\) & \(55.00\%\) \\
AugSmolVLA & \(\mathbf{56.67\%}\) & \(\mathbf{55.00\%}\) & \(\mathbf{72.50\%}\) & \(\mathbf{75.00\%}\) & \(\mathbf{56.67\%}\) & \(\mathbf{58.33\%}\) \\
\bottomrule
\end{tabular*}
\end{table*}

\begin{table*}[!t]
\caption{Comparison of dual-arm task experimental results.}
\label{tab5}
\centering
\small
\renewcommand{\arraystretch}{1.25}
\begin{tabular*}{\textwidth}{@{\extracolsep{\fill}}lcccc@{}}
\toprule
\textbf{Model}
& \makecell{\textbf{Unscrew Tube Cap}\\\textbf{(SR)}} 
& \makecell{\textbf{Tighten Tube Cap}\\\textbf{(SR)}} 
& \makecell{\textbf{Pour Waste Liquid}\\\textbf{(SR)}} 
& \makecell{\textbf{Composite task}\\\textbf{(CR)}} \\
\midrule
Act        & \(11.67\% \pm 1.67\) & \(8.33\% \pm 4.41\)  & \(3.33\% \pm 1.67\)  & \(7.65\%\)  \\
X-vla      & \(5.00\% \pm 2.89\)  & \(11.67\% \pm 1.67\) & \(10.00\% \pm 2.89\) & \(11.26\%\) \\
Smolvla    & \(51.67\% \pm 4.41\) & \(28.33\% \pm 4.41\) & \(33.33\% \pm 6.01\) & \(37.00\%\) \\
AugSmolVLA & \(\mathbf{55.00\% \pm 5.77}\) & \(\mathbf{33.33\% \pm 3.33}\) & \(\mathbf{36.67\% \pm 3.33}\) & \(\mathbf{39.45\%}\) \\
\bottomrule
\end{tabular*}
\end{table*}

\begin{figure*}[!t]
  \centering
  \includegraphics[width=0.7\textwidth]{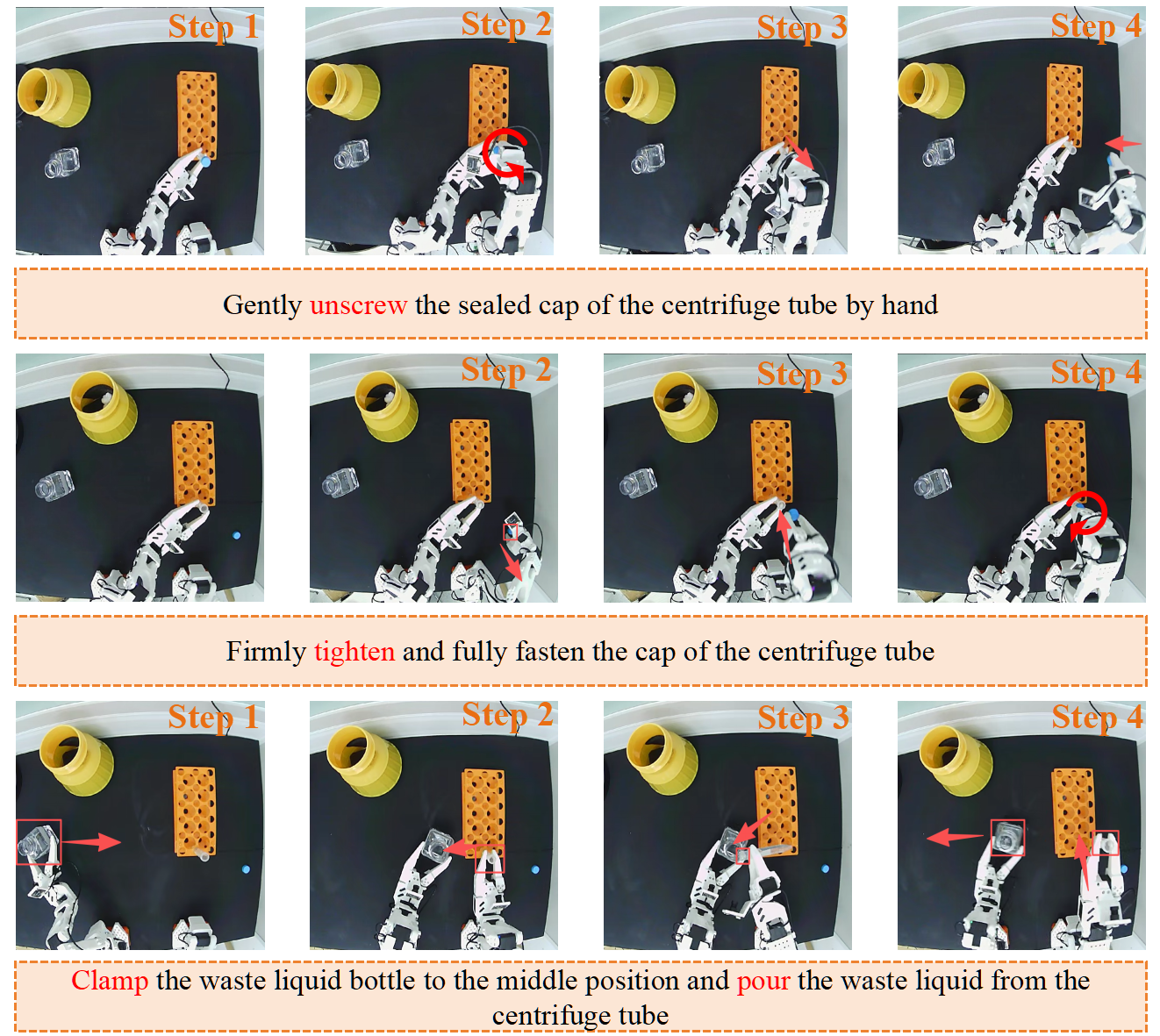}
  \caption{Examples of Dual-Arm task execution processes.}
  \label{fig7}
\end{figure*}

These results indicate that AugSmolVLA maintains better execution stability in multi-step biological manipulation. Since composite tasks require continuous state preservation across atomic operations, early errors can easily propagate to subsequent steps. The performance improvement is mainly attributed to the visual perturbation data augmentation strategy, which enhances the model’s adaptability to illumination changes, local reflections, and transparent labware interference, thereby improving object localization and action stability during task transitions.

\subsection{Dual-Arm Task Experiments}

\begin{table*}[!t]
\caption{Comparison of experimental results under the overexposure condition.}
\label{tab6}
\centering
\scriptsize
\renewcommand{\arraystretch}{1.25}
\begin{tabular*}{\textwidth}{@{\extracolsep{\fill}}lccccccc@{}}
\toprule
\textbf{Model}
& \makecell{\textbf{Place Centrifuge}\\\textbf{Tube to Orange}\\\textbf{Rack (SR)}}
& \makecell{\textbf{Place Cryotube to}\\\textbf{Red Rack (SR)}}
& \makecell{\textbf{Discard}\\\textbf{Centrifuge}\\\textbf{Tube (SR)}}
& \makecell{\textbf{Discard}\\\textbf{Cryotube (SR)}}
& \makecell{\textbf{Pour Waste}\\\textbf{Liquid}}
& \makecell{\textbf{Tighten Tube}\\\textbf{Cap}}
& \makecell{\textbf{Loading}\\\textbf{centrifuge}\\\textbf{tube (CR)}} \\
\midrule
Normal-smolvla
& \(43.33\% \pm 3.33\)
& \(40.00\% \pm 2.89\)
& \(88.33\% \pm 1.67\)
& \(48.33\% \pm 1.67\)
& \(51.67\% \pm 4.41\)
& \(28.33\% \pm 4.41\)
& \(55.00\%\) \\

Exposure-smolvla
& \(33.33\% \pm 7.27\)
& \(31.67\% \pm 6.01\)
& \(71.67\% \pm 1.67\)
& \(31.67\% \pm 4.41\)
& \(40.00\% \pm 0.00\)
& \(11.67\% \pm 4.41\)
& \(39.60\%\) \\

AugSmolVLA
& \(\mathbf{43.33\% \pm 4.41}\)
& \(\mathbf{71.67\% \pm 6.01}\)
& \(\mathbf{78.33\% \pm 4.41}\)
& \(\mathbf{43.33\% \pm 4.41}\)
& \(\mathbf{48.33\% \pm 1.67}\)
& \(\mathbf{16.67\% \pm 1.67}\)
& \(\mathbf{44.00\%}\) \\
\bottomrule
\end{tabular*}
\end{table*}

Following the composite-task experiments that evaluated multi-step continuous execution, we further conducted dual-Arm task experiments to assess the performance of the proposed method in coordinated multi-manipulator operation scenarios. Compared with single-arm tasks, dual-Arm Tasks involve more complex temporal dependencies and impose higher requirements on spatial coordination and action consistency between the two arms. Therefore, representative dual-Arm operations, including tube-cap twisting and liquid pouring, were selected for comparative evaluation under the same experimental settings. The experimental results are shown in Table 5, and corresponding examples of the task execution process are presented in Fig \ref{fig7}.

Overall, all methods achieved markedly lower success rates in dual-Arm Tasks than in single-arm scenarios, indicating that these tasks impose higher control requirements on the models. In particular, the twisting tasks, including Unscrew Tube Cap and Tighten Tube Cap, as well as the pouring task, Pour Waste Liquid, require precise coordination and force-constrained control between the two arms, making them sensitive to temporal synchronization and contact stability. In these tasks, the proposed method showed clear advantages. For example, it achieved 55.00\% on Unscrew Tube Cap, exceeding ACT (11.67\%) by more than 40 percentage points. In Pour Waste Liquid, it reached 36.67\%, clearly outperforming X-VLA (10.00\%) and ACT (3.33\%). In addition, for the overall composite task, AugSmolVLA achieved 39.45\%, showing a further improvement over SmolVLA (37.00\%) and the other baseline methods.

\subsection{Ablation Experiments under Overexposed Conditions}
Based on the preceding experiments under normal lighting conditions, we further constructed a overexposure scenario for comparative evaluation to assess model robustness under severe visual disturbances. The overexposure environment was created by adjusting the camera exposure parameters, introducing obvious overexposure and loss of visual details. Under this setting, three model variants were compared: SmolVLA trained and tested under normal conditions, which served as the baseline; SmolVLA directly tested under the overexposure condition, which was used to evaluate performance degradation caused by environmental changes; and AugSmolVLA tested under the overexposure condition, which was used to verify the effect of data augmentation on model generalization. The experimental results are shown in Table \ref{tab6}.

The overexposed setting imposed a pronounced visual distribution shift on SmolVLA, directly weakening its ability to localize transparent labware and maintain stable manipulation in visually degraded scenes. Without data augmentation, this degradation was reflected in consistent performance drops across precise placement, disposal, and multi-step manipulation tasks. For example, the success rate decreased from 40.00\% to 31.67\% in Place Cryotube to Red Rack, from 48.33\% to 31.67\% in Discard Cryotube, and from 55.00\% to 39.60\% in the composite task Loading centrifuge tube. These results indicate that visual information degradation caused by overexposure can impair the model’s ability to recognize and localize target objects. In contrast, AugSmolVLA achieved clear performance recovery or improvement in most tasks. For example, the success rate of Place Cryotube to Red Rack increased from 31.67\% to 71.67\%, that of Discard Cryotube increased from 31.67\% to 43.33\%, and that of Place Centrifuge Tube to Orange Rack recovered to 43.33\%, which was comparable to its performance under normal lighting. Overall, data augmentation effectively alleviated the distribution shift under overexposure conditions, enabling the model to maintain more stable manipulation performance when visual information was degraded and demonstrating stronger environmental robustness.

\section{Conclusion}
This study presents BioProVLA-Agent, a protocol-grounded multi-agent framework for verifiable robotic biological experimentation. Unlike conventional robotic manipulation pipelines that primarily rely on direct instruction-to-action mapping, the proposed framework establishes a closed-loop workflow that integrates biological protocol understanding, structured task decomposition, VLM-RAG-based state verification, and VLA-driven embodied execution. Through the collaboration of the Guiding Decision Agent, Specialized LLM Protocol Agent, VLM-RAG Verification Agent, and VLA Embodied Agent, unstructured biological protocols are transformed into executable subtask sequences with explicit action instructions, preconditions, completion criteria, and knowledge-based indices. This design enables the robotic system not only to execute experimental operations, but also to verify task feasibility before execution, assess completion states after execution, and provide interpretable feedback for long-horizon biological workflows.

In addition, AugSmolVLA is developed by incorporating a visual perturbation data augmentation strategy into SmolVLA, improving the robustness of VLA-based execution under illumination changes, local reflections, transparent labware interference, and overexposed visual conditions. A biological robotic manipulation benchmark is further constructed on the So-ARM101 platform, covering 15 atomic tasks, single-arm operations, composite tasks, dual-arm coordination, and overexposure robustness evaluation. The experimental results demonstrate that the proposed method improves execution stability across different biological manipulation scenarios. For example, AugSmolVLA increased the success rate of Place Cryotube to Red Rack from 40.00\% to 65.00\% in single-task experiments and achieved a completion rate of 75.00\% in the composite task Clean up waste materials, exceeding the original SmolVLA by 12.50 percentage points. Under overexposed conditions, it further improved the success rate of Place Cryotube to Red Rack from 31.67\% to 71.67\%, demonstrating stronger robustness against severe visual disturbance. These results indicate that integrating protocol-grounded reasoning, retrieval-augmented visual verification, and robust VLA execution provides an effective technical pathway toward reliable robotic biologists for real wet-lab automation.

This study still has several limitations. The current experiments mainly focus on basic biological operations and limited scenarios, while task complexity, protocol length, and object diversity remain to be further expanded. In addition, Dual-Arm coordination tasks still pose considerable challenges in fine contact control and temporal coordination. Future work will build upon this framework by incorporating dexterous hands and multimodal perception to support more complex grasping, twisting, pipetting, and sample-handling tasks. Furthermore, experimental design, result analysis, and autonomous decision-making modules will be integrated to advance the system from “executing experimental steps” toward a robotic biologist capable of autonomously conducting experiments and assisting scientific discovery.

\section*{Funding Statement}
This work was supported by the National Natural Science Foundation of China (Grant No. 62476087) and the Shanghai Municipal Education Commission Initiative on Artificial Intelligence-Driven Reform of Scientific Research Paradigms and Empowerment of Discipline Leapfrogging.

\section*{Code availability}
Our code and video demonstrations are publicly available at: https://github.com/no-guess/BioProVLA-Agent.

\nocite{*}
\bibliography{references}

\end{document}